\documentclass{article}

\usepackage{arxiv}

\usepackage[utf8]{inputenc} 
\usepackage[T1]{fontenc}    
\usepackage{hyperref}       
\usepackage{url}            
\usepackage{booktabs}       
\usepackage{amsfonts}       
\usepackage{nicefrac}       
\usepackage{microtype}      
\usepackage{tikz}
\usepackage{lipsum}
\usepackage{graphicx}
\graphicspath{ {./images/} }
\usepackage{pgfplotstable}
    \pgfplotsset{
        compat=1.9,
        compat/bar nodes=1.8,
    }
    \pgfplotstableread{
        Label      Centroid     NMS   Harris FAST 
        720$\times$540    24.1     7.3   27.2  41.4       
        1080$\times$720    21.3 8.5     32.2    38     
        1440$\times$1080   22.9 8.5     32.3    36.3
        1920$\times$1080   23.8 8.1     31.8    36.3
    }\testdata

\usepackage{xcolor}
\usepackage{algorithm}
\usepackage{setspace}
\usepackage{amsmath}
\usepackage{multirow}
\usepackage{algpseudocode}
\usepackage{threeparttable}
\usepackage{pgfplots}
\usepackage{caption}
\usepackage{subcaption}

\usetikzlibrary{patterns}

\definecolor{color1}{HTML}{3594cc} 
\definecolor{color2}{HTML}{ea801c} 
\definecolor{color3}{HTML}{8cc5e3} 
\definecolor{color4}{HTML}{f0b077} 
\definecolor{color5}{HTML}{00b0be} 
\definecolor{color6}{HTML}{8fd7d7} 
\definecolor{color7}{HTML}{eaa91c}
\definecolor{color8}{HTML}{cf761c} 

\title{Faster than Fast: Accelerating Oriented FAST Feature Detection on Low-end
  Embedded GPUs}

\author{
 Qiong Chang \\
  School of Coumputing\\
  Institute of Science Tokyo\\
  Tokyo, Japan 152-8550 \\
  \texttt{q.chang@c.titech.ac.jp} \\
   \And
 Xinyuan Chen \\
  Dalian University of Technology\\
  Dalian, China 210093  \\
  \And
 Xiang Li \\
  Nanjing University\\
  Nanjing, China. 210093 \\
  \And
 Weimin Wang*\\
  Dalian University of Technology\\
  Dalian, China 210093  \\
  \texttt{wangweimin@dlut.edu.cn} \\
  \And
 Jun Miyazaki \\
  School of Computing\\
  Institute of Science Tokyo\\
  Tokyo, Japan 152-8550 \\
  \texttt{miyazaki@c.titech.ac.jp} \\
}

\begin{document}
\maketitle
\begin{abstract}
The visual-based SLAM (Simultaneous Localization and Mapping) is a technology
widely used in applications such as robotic navigation and virtual reality,
which primarily focuses on detecting feature points from visual images to
construct an unknown environmental map and simultaneously determines its own
location. It usually imposes stringent requirements on hardware power
consumption, processing speed and accuracy. Currently, the ORB (Oriented FAST and
Rotated BRIEF)-based SLAM systems have exhibited superior performance in terms
of processing speed and robustness. However, they still fall short of meeting the
demands for real-time processing on mobile platforms. This limitation is
primarily due to the time-consuming Oriented FAST calculations accounting for
approximately half of the entire SLAM system. This paper presents two methods to
accelerate the Oriented FAST feature detection on low-end embedded GPUs. These
methods optimize the most time-consuming steps in Oriented FAST feature
detection: FAST feature point detection and Harris corner detection, which is
achieved by implementing a binary-level encoding strategy to determine candidate
points quickly and a separable Harris detection strategy with efficient
low-level GPU hardware-specific instructions. Extensive experiments on a Jetson
TX2 embedded GPU demonstrate an average speedup of over 7.3 times compared to
widely used OpenCV with GPU support. This significant improvement highlights
its effectiveness and potential for real-time applications in mobile and
resource-constrained environments.
\end{abstract}

\section{Introduction}
SLAM (Simultaneous Localization and Mapping)~\cite{SLAM} is an algorithm that
constructs an unknown environmental map and simultaneously determines its own
location, primarily used in robotics and self-driving
applications~\cite{ben2022edge}. Various sensor modes, such as sonar,
LiDAR~\cite{he2024robust}, and cameras, are employed in different SLAM
systems. Among these, Feature-Based Visual SLAM (VSLAM)~\cite{azzam2020feature}
is particularly convenient and cost-effective, leading to extensive research in
this area. As feature-based VSLAM has advanced, numerous feature point
extraction methods have been proposed, including
SIFT~\cite{lowe2004distinctive}, SURF~\cite{bay2008speeded},
FAST~\cite{viswanathan2009features}, and even learning-based
methods~\cite{xu2020cnn}~\cite{dusmanu2019d2}. Most current mainstream SLAM
systems are deployed in mobile applications, which impose stringent requirements
on real-time performance, computational resources, and power consumption. While
SIFT, SURF, and learning-based descriptors account for changes in lighting,
scale, and other factors during image transformations and can extract precise
feature points, they are computationally expensive and often fail to meet the
performance needs of SLAM systems. On the other hand, ORB (Oriented FAST and
Rotated BRIEF)~\cite{rublee2011orb}, a lightweight feature description method,
addresses these challenges by using Oriented FAST (Features from accelerated
segment test) for feature point detection and rotated BRIEF (Binary Robust
Independent Elementary Features) for feature description, trading off some
accuracy and robustness for improved computational speed.
Currently, ORB has become the predominant feature point detection method used in
SLAM systems. However, with the continuous improvement in image resolution,
ORB-based SLAM systems still struggle to meet real-time processing requirements
on most mobile platforms. The primary reason is that to enhance accuracy, the
Oriented FAST detection step of modern ORB algorithms typically requires
constructing multi-level pyramids to detect features at various
scales. Additionally, it also involves matching Harris corner points for each
candidate feature point identified by the FAST algorithm to screen more stable and robust
feature points. Although the computation of the FAST algorithm itself is
straightforward, it is highly repetitive and the processing of adjacent pixels
is locally independent, making it challenging to further optimize the
algorithm's complexity. Therefore, Oriented FAST feature point detection typically accounts for approximately half of the computing time in ORB, which itself constitutes over 65\% of the total computation in SLAM systems~\cite{orb-slam3}.
To enhance the practicality of SLAM systems, numerous researchers are exploring
methods to accelerate Oriented FAST processing on mobile platforms. The primary
method involves leveraging high-performance acceleration hardware such as
FPGAs and GPUs. Researchers, including those cited
in~\cite{taranco2023locator,liu2019eslam,sun2019flexible,vemulapati2022fslam},
have successfully built pipelines on FPGAs and designed reusable on-chip BRAM to
store intermediate results, thereby speeding up Oriented FAST
processing. Although using FPGAs for large-scale SLAM systems offers certain
advantages in terms of energy efficiency, it comes with high device costs and
long development cycles. More critically, customized hardware designs struggle
to maintain consistent SLAM system performance under varying conditions. In
contrast, embedded GPUs have become the preferred choice for high-performance
mobile intelligent systems due to their superior parallel processing
capabilities and more flexible software computation methods~\cite{Chang_TinyStereo}. Mimura et
al.~\cite{mimura2022acceleration} proposed an asynchronous processing method for
object detection using an embedded GPU. Their method leverages the CPU to
handle object detection tasks, thereby offsetting the GPU overhead during
Oriented FAST feature point detection. This method enabled real-time detection
for a single-layer ORB system, achieving a performance of nearly 81 FPS on a
Jetson Nano GPU. Muzzini et al.~\cite{muzzini2023brief} employed GPU
multi-channel stream processing to decompose the ORB process and balance the
load of GPU computing resources, thereby accelerating the entire system's
processing. Their methods significantly improve the execution efficiency of the
Oriented FAST algorithm on a Jetson AGX Xavier GPU. Zhi et
al.\cite{zhi2019realization} accelerated the FAST detection by leveraging the
parallel processing capabilities of the Tesla K40c GPU architecture by CUDA. They allocated GPU
resources equally to each pixel, enabling the efficient processing of FAST
feature detection task for each pyramid level. This method achieved a
processing speed 6\textasciitilde10 times faster than that of a CPU. All methods aim to
leverage the GPU to enhance the overall ORB processing flow; however, they do
not significantly optimize the Oriented FAST algorithm itself. Recently, a fast
framework of ORB system (CUDA\_ORB) has been published by~\cite{cuda-ORB}. They
addressed the limitations of library functions by implementing the ORB algorithm
with pixel-level processing on GPUs. Their performed the detection over 3,000
feature points in a four-layer pyramid within a 1920$\times$1080 image,
achieving a processing speed of 57fps on a Jetson TX2 GPU, which is faster than the GPU-based OpenCV library. However, there is still room for
improvement in terms of feature point detection and memory usage. An analysis of their runtime distribution for each step across various image sizes is provided in Fig~\ref{fig1}. The results indicate that FAST and Harris detection are the most
time-consuming steps in Oriented FAST algorithm. 
Specifically, FAST feature detection involves numerous branch instructions, which in parallel execution can be several to dozens of times less efficient than basic arithmetic operations like addition, subtraction, and multiplication. Furthermore, Harris detection requires frequent access to randomly located features, leading to significant overhead on embedded GPUs due to their high memory access latency.
Therefore, to further enhance the
performance of the Oriented FAST and facilitate its application in SLAM systems, this paper introduces significant optimizations, including:
\begin{figure}
\center
\begin{tikzpicture}
    \begin{axis}[
        ybar stacked,
        bar width=20pt,
        ymin=0,
        ymax=100,
        width=3.5in,
        height=2in,
        xtick=data,
        ylabel={Runtime Distribution (\%)},
        xlabel={Image Size},
        font=\footnotesize,
        legend style={
            font=\tiny,
            cells={anchor=west},
            at={(1.25,1)},
        },
        reverse legend=true,
        xticklabels from table={\testdata}{Label},
    ]
        \addplot [fill=color2]
            table [y=FAST, meta=Label, x expr=\coordindex]
                {\testdata};
                \addlegendentry{FAST}
        \addplot [fill=color5]
            table [y=Harris, meta=Label, x expr=\coordindex]
                {\testdata};
                    \addlegendentry{Harris}
        \addplot [fill=color1]
            table [y=NMS, meta=Label, x expr=\coordindex]
                {\testdata};
                \addlegendentry{NMS}
        \addplot [fill=color3]
            table [y=Centroid, meta=Label, x expr=\coordindex]
                {\testdata};
                    \addlegendentry{Centroid}
    \end{axis}
\end{tikzpicture}
\caption{Runtime distribution of each step in Oriented-FAST using a Jetson TX2 GPU}
\label{fig1}
\end{figure}
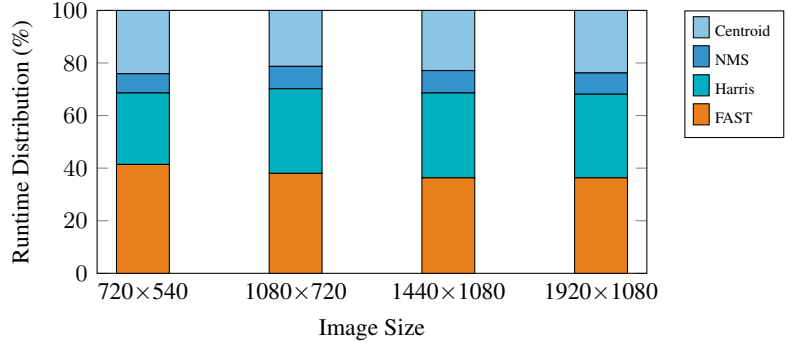

\begin{itemize}
\item An optimized FAST detection strategy is proposed, employing a binary encoding strategy to quickly identify candidate feature points, significantly reducing the number of branch statements. PTX code~\cite{ptx} analysis shows that this strategy saves over 35\% of global memory loads and branch statements, achieving nearly a roughly 1.2x speedup.

\item A semi-separable Sobel operator is proposed which can effectively
  accelerate the Harris detection by utilizing a circular buffer on fast but size-constrained shared memory. This method significantly improves the detection speed by an average of 7.3 times.

\item Shared memory is leveraged to integrate FAST and Harris steps, minimizing the transmission cost for pixels and feature points data.
\end{itemize}
Additionally, extensive experiments are conducted to evaluate each optimization scheme presented in this paper. The proposed method significantly improved the processing speed of Oriented FAST compared to other methods, including the most popular
OpenCV libraries across various images.
This paper is organized into the following sections:
Section~\ref{sec:relatedwork} reviews related
work. Section~\ref{sec:background} provides background on the FAST and Harris
detection algorithms and the embedded GPU architecture. Section~\ref{sec:analysis} analyzes the current bottlenecks in implementing Oriented FAST. Section~\ref{sec:optimization} details the optimization methods, and Section~\ref{sec:experiment} presents the experimental results. Finally, Section~\ref{sec:conclusion} concludes the paper.

\section{Related work}
\label{sec:relatedwork}
\subsection{Acceleration of FAST feature point detection}
Nagy et al.\cite{nagy2020faster} develop a novel method for non-maximum
suppression specifically tailored for GPU architectures, allowing for
simultaneous selection and extraction of local response maximum and spatial
feature distribution. It simplifies the process of the original FAST detection
method by eliminating the distinction between dark and bright points when
assessing feature points. Then, the FAST feature classification can be significantly
streamlined by utilizing a 16-bit array to store comparison features and
pre-establishing an 8KB lookup table. This method effectively reduces
computational complexity and improves the efficiency of feature detection.

Muzzini et al.\cite{muzzini2023brief} present a novel method to enhance
ORB-SLAM's performance by implementing a parallel GPU-based solution
specifically for the tracking component. Unlike traditional GPU ports, this
method introduces an innovative technique for constructing image pyramids
directly on the GPU, significantly reducing computational overhead. Each pixel
at each level is assigned to a CUDA GPU thread, which calculates its value from
the original image, eliminating dependencies between pyramid levels and reducing
memory copying between the CPU and GPU.

Park et al.\cite{jun2012a} presents a high-performance hardware accelerator
tailored for embedded vision applications. This accelerator enhances the FAST
algorithm by optimizing joint algorithm-architecture for bit-level parallelism,
leading to significant performance improvements. The system achieves a 9.5x
performance boost compared to state-of-the-art processors while only using 30\%
of the logic gates, thanks to its low-power unified hardware platform. With a
throughput of 94.3 frames per second in Full HD resolution at a power
consumption of 182 mW, it ensures efficient real-time image recognition suitable
for mobile devices and vehicles.

\subsection{Acceleration of Harris corner detection}
The Harris corner detection uses Sobel operator masks to
calculate image derivatives, angle responses, and suppresses non maximum points
to obtain feature points. 
He et al.\cite{he2018parallel} accelerates  the Harris algorithm by implementing
a  parallel method  using OpenCL  on a  heterogeneous architecture.  The authors
optimize  the many-core  processor  by distributing  the computational  workload
across both the CPU and GPU.  They combine the gaussian blur convolution, image
gradient and Harris  matrix into one GPU kernel, significantly  reduces the data
transmission and  enhances the  performance of the  Harris algorithm,  making it
more  suitable  for  real-time  processing on  devices  with  limited  computing
resources.  The experimental  results  demonstrate  substantial improvements  in
processing  speed  and  efficiency,   which  is  critical  for  high-performance
computing applications involving real-time image and video processing.
Haggui et al.\cite{haggui2018harris} explores the optimization of the Harris
corner detection algorithm for Non-Uniform Memory Access (NUMA) architectures to
enhance its efficiency and scalability. The authors focus on the inherent
stencil-based data access patterns of the Harris algorithm, which require
meticulous memory management to minimize cache misses and leverage the NUMA
architecture effectively. They implement a SIMD (Single Instruction, Multiple
Data) version of the algorithm to exploit wide vector registers, ensuring better
parallel processing capabilities. By directly and explicitly incorporating
common and novel optimization strategies, it demonstrates significant
improvements in scalability and performance on a dual-socket Intel
Broadwell-E/EP system, making it more suitable for real-time applications.
Loundagin\cite{loundagin2015optimizing} optimizes the
implementation of Harris detection on an NVIDIA GPU. The implementation loads
the filter mask into the constant memory to optimize the memory access and
creates an integration image to perform the gradient product of Harris score
which optimizes the ROI total calculation. Thus, regardless of the neighborhood
dimension, calculating the sum of neighborhoods only requires four global memory
accesses, significantly improves the efficiency of calculating Harris score for
feature points.

\section{Background}
\label{sec:background}
\subsection{Features from accelerated segment test (FAST)}
\begin{figure}
    \centering
    \includegraphics[width=0.8\linewidth]{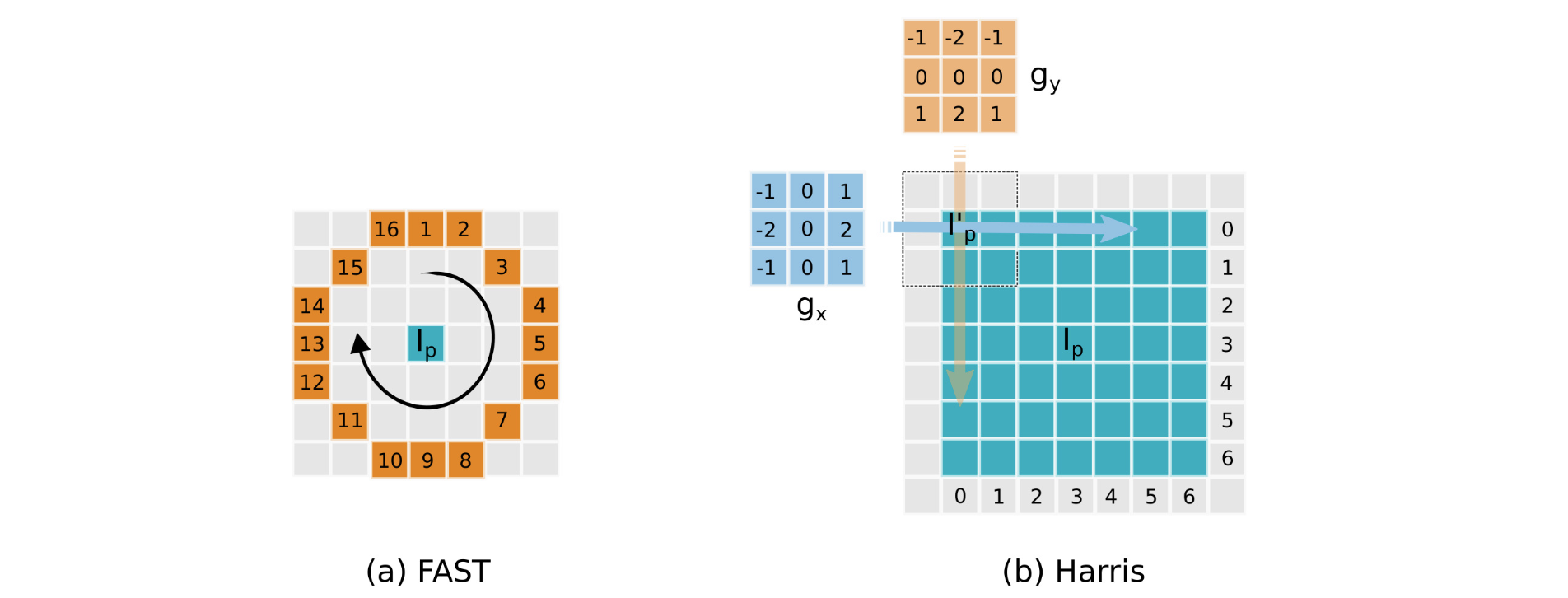}
    \caption{Two primary detection steps in Oriented FAST. (a) An example
      of FAST feature point detection circle, including a candidate pixel and its 16
      surrounding pixels. (b) An example of Harris corner detection, using the
      Sobel operator to calculate the gradients in two directions.}
    \label{fig2}
\end{figure}

The Features from Accelerated Segment Test (FAST) is a high-performance feature
point detection algorithm used in many computer vision applications. It involves
a simple decision tree to classify a pixel as a feature point based on the intensity
differences $D_t$ with its surrounding pixels. Specifically, as
Fig~\ref{fig2}(a) shows, it examines a circular region of sixteen pixels
$I_{i}$ around a candidate pixel $I_p$:
\begin{equation}
    D_{t}(I_p, I_i)=\left\{\begin{matrix} dark\hfill\qquad I_{i}\le I_{p}-t,
    \hfill \\similar \hfill\qquad I_{p}-t <  I_{i}<I_{p}+t,
    \hfill \\bright\hfill\qquad I_{i}\ge I_{p}+t.
\hfill\end{matrix}\right.
\end{equation}
It compares the difference in pixels with the threshold value t and classifies the surrounding 16 pixels into three types based on different comparison results: dark, similar, and bright. When there are 9 (or more) continuous surrounding pixels that are in a
dark or bright state, the candidate point $I_p$ is defined as a feature
point. This process allows FAST to quickly identify key features in an image,
which are crucial for various computer vision tasks.

\subsection{Harris corner detector}
Due to FAST often producing a large number of feature points locally, Harris is used to further filter these points detected by FAST, allowing the
selection of the most robust and reliable feature points. The Harris corner
detector operates by analyzing the local auto-correlation function of a signal,
which measures the changes in intensity within a small window. As shown in
Fig~\ref{fig2}(b), by computing the gradient $g_x$ and $g_y$ of the image with
the Sobel operator, a second-moment matrix $M$ consisting of $g_x$ and $g_y$ can
be obtained by:
\begin{equation}
  M=\begin{bmatrix}
        g_x^2 & g_xg_y\nonumber\\
        g_xg_y & g_y^2\\
  \end{bmatrix}
\end{equation}
where
\begin{equation}
  g_x=\frac{\partial{I}}{\partial{x}},\ \ \ \ \  g_y=\frac{\partial{I}}{\partial{y}}.
\end{equation}
Then, the Harris score function can be calculated by examining the
eigenvalues $det(M)$ and the traces of $trace(M)$ as follow:
\begin{equation}
Score=det(M)-k\cdot(trace(M))^2,\nonumber
\end{equation}
where 
\begin{equation}
 det(M)=g_x^2g_y^2-(g_xg_y)^2,\ \ \ \ \ trace(M)=g_x^2+g_y^2.
\label{equ3}
\end{equation}
$k$ is an empirical constant in range of (0.04, 0.06). In CUDA\_ORB and Opencv library, $k$ is set to a constant of 0.04.
The Harris score identifies regions where there is a significant change in
intensity in multiple directions. These regions are marked as corners. The
Harris corner detector is particularly effective because it is invariant to
rotation, scale, and illumination changes, making it a robust choice for tasks
such as image matching, motion tracking, and 3D reconstruction.

\subsection{Architecture of Jetson Embedded GPUs}
Jetson embedded GPUs are designed specifically for edge applications, combining
the efficiency of ARM CPUs with powerful CUDA-capable GPUs in a compact form
factor. The architecture typically includes an ARM-based CPU, an NVIDIA GPU with
CUDA cores, and various interfaces for sensors and peripherals, all optimized
for low power consumption. Unlike high-end GPUs, which prioritize raw
performance for PCs and workstations, Jetson GPUs focus on energy efficiency and
integration for embedded systems in constrained power and space environments. A
key architectural difference is the use of LPDDR4/LPDDR5 memory with a bandwidth
of approximately 50 GB/s, about one-tenth that of general-purpose
GPUs. Therefore, optimizing algorithms for Jetson GPUs requires minimizing
global memory access and maximizing the reuse of on-chip memory.

\section{Performance Bottleneck Analysis}
\label{sec:analysis}
In this section, a detailed analysis of the common challenges encountered when using GPUs for FAST and Harris detection is provided, along with the corresponding optimization methods.

\begin{algorithm}[t]
\setstretch{1.1} 
\footnotesize
\caption{A pseudo code of FAST feature point detection in~\cite{cuda-ORB} }
\begin{algorithmic}[1]
\State \textbf{Input}: $I_p$, $I_i[16]$ \Comment{Candidate pixel $I_p$ and its surrouding 16 pixels $I_i$.}
\State \textbf{Output}: $isPoint$  
\State $isPoint \leftarrow False$; \Comment{Initialization.}
\State $D_{flag} \leftarrow checkD_t(I_p, I_i[16])$ \Comment{Check 8 pairs of $D_t$ with 16 pixels $I_i$ and set a unique tone}.
  \If{$D_{flag}$ is dark}
     \For{$k=0$ to 25}\Comment{16+9 classifications are required to determine the $D_t$ consistency within 16 segments.}
        \State $count \leftarrow 0$;
         \If{$D_t\geq{t}$}
         \If{$count++>9$}
         \State $isPoint \leftarrow True$;
         \State break; \Comment{Skip the remaining classifications.}
         \EndIf
         \Else
          \State $count \leftarrow 0$;
         \EndIf
     \EndFor
  \Else \Comment{$D_{flag}$ is bright.}
     \For{$k=0$ to 25}  \Comment{16+9 classifications are required to determine the $D_t$ consistency within 16 segments.}
     \State $count \leftarrow 0$;
       \If{$D_t\leq{-t}$}
         \If{$count++>9$}
         \State $isPoint \leftarrow True$;
         \State break; \Comment{Skip the remaining classifications}.
         \EndIf
         \Else
         \State $count \leftarrow 0$;
         \EndIf
     \EndFor
  \EndIf
\State \Return $isPoint$;
\end{algorithmic}
\label{algo1}
\end{algorithm}

\subsection{Analysis of FAST feature point detection}
First, as illustrated in Fig~\ref{fig2}(a), the process of FAST feature point
  detection involves repeatedly computing the pixel intensity difference $D_t$
  between a central candidate pixel $I_p$ and its surrounding 16 pixels $I_i$ to
  extract feature points. This ring-shaped pixel configuration lacks the memory
  continuity found in a rectangular window and maintains a fixed position,
  leading to inefficiencies. The calculations between adjacent central candidate
  pixels cannot be reused, even though the difference $D_t$ of each pixel pair
  could theoretically be shared by two sets of candidate points. This sharing
  mechanism, however, results in significant memory consumption and requires
  complex memory indexing, failing to simplify the overall computational
  process. Therefore, this intrinsic structure of the FAST algorithm poses
  challenges for further reduction of computational complexity through
  algorithmic optimization. The need for continuous pixel difference
  calculations, coupled with non-reusable intermediate computations and
  inefficient memory usage, underscores the difficulty in optimizing the FAST
  feature detection algorithm for improved performance and reduced resource
  consumption.
  Second, to check whether there are 9 continuous $D_t$ values that meet the conditions for a candidate
  pixel $I_p$, a maximum of 16 segments need to be checked for consistency. This
  operation generally has a high degree of parallelism, but brute force
  computation is not an appropriate choice. The approach in \cite{cuda-ORB} employs a strategy
  shown to reduce calculations, as shown in Algorithm~\ref{algo1}. It first
  checks the consistency between the beginning and the end of each segment,
  such as $D_t(I_p, I_1)$ and $D_t(I_p, I_9)$ (line 4), to preliminarily judge
  whether the overall state $D_{flag}$ is bright or dark. Then, the continuity
  within each segment is further checked with a loop based on the value of
  $D_{flag}$ (line 6 to 16 and line 18 to 28). The loop exits immediately upon
  finding any segment with 9 continuous $D_t$ values that meet the conditions for a candidate
  pixel $I_p$, minimizing unnecessary
  operations (line 11 and line 23). This process continues until a consecutive
  segment is found or all 16 segments have been checked. Although this strategy
  eliminates redundant calculations during the consistency determination, its
  execution efficiency on the GPU is suboptimal. Firstly, checking the
  consistency of $D_t$ requires numerous branch instructions {\em if-else}, which
  significantly reduce throughput across all instruction sets. Secondly, because
  GPU threads execute synchronously within a {\em Warp}, the premature
  termination of any thread not only fails to enhance execution efficiency but
  also causes a reduction due to idle waiting and increases the risk of
  out-of-order memory access. Therefore, this kind of method that reduces
  computation through prejudgment cannot effectively improve the execution
  efficiency of FAST detection on a GPU.

\subsection{Analysis of Harris corner detection}
\label{analysis:harris}
  First, as mentioned earlier, Harris detection aims to eliminate redundant feature
  points detected by FAST, meaning its computation relies on FAST detection
  results. Because their computations are relatively independent and their
  kernels are usually designed separately, most current SLAM systems execute
  them in a step-by-step manner. When driving a GPU to boot a SLAM system, FAST
  feature points are typically stored in global memory first, followed by the
  Harris kernel loading these feature points and their surrounding pixel
  information to complete the corner detection. While this method is easy to
  implement, it poses significant performance limitations for embedded GPUs with
  limited memory bandwidth: 1) the load and store operations for FAST
  feature points and related pixel information increase memory access demands,
  leading to high latency that can severely impact the processing speed of the
  SLAM system; 2) the random distribution of FAST feature points greatly
  increases the risk of discontinuous memory access, which significantly
  impacts the performance of the entire GPU-based SLAM systems.
  Second, Harris score calculation primarily involves applying the 3$\times${3}
  Sobel operator filter to each pixel within a 7$\times$7 window around a FAST feature
  point. This operation, characterized by high parallelism and substantial
  computational load, is well-suited for leveraging the GPU's multi-threading
  capabilities to enhance computation speed. However, the random positions and
  varying quantities of FAST feature points across different input images
  present a challenge for evenly allocating GPU threads and memory resources for
  the Harris score calculation of each feature point. Consequently, most current
  methods resort to using a single thread to perform the calculation for each
  FAST feature point, which leads to significant GPU resource wastage.
  Finally, driving a single thread to perform Harris score calculation for each FAST
  feature point helps to allocate hardware resources in a targeted
  manner. However, applying a 3$\times$3 Sobel operator on a 7$\times$7 window
  for serial filtering operations is highly inefficient, causing active threads
  to be busy while inactive threads remain idle. This causes the Harris computation to become the bottleneck for the entire Oriented FAST process.


\section{Optimization Methodology}
\label{sec:optimization}
After analyzing the bottlenecks in Oriented FAST processing, a series of optimization methods is proposed for FAST and Harris detection on GPUs. 
For FAST, a strategy is introduced that encodes the 16 differences $D_t$ into a 32-bit unsigned integer to enhance consistency checks on the GPU. For Harris, it is combined with FAST to reduce the data transmission cost, and additional strategies are developed to improve the execution speed of the Sobel operator.

\subsection{Optimization for FAST feature point detection}
\begin{figure}[t]
    \centering \includegraphics[width=1\linewidth]{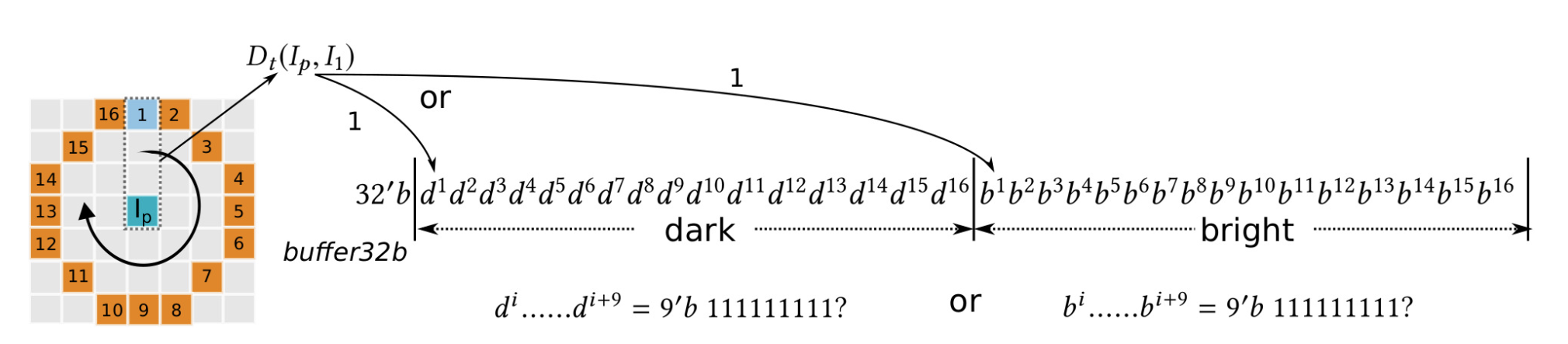}
    \caption{Rapid consistency determination using a binary encoding
      strategy. 16 differences $D_t$ are stored as Boolean values within a
      32-bit binary number, divided into two groups representing their bright
      and dark states. The FAST feature point can be determined by verifying the
      binary values of any continuous 9-bit segment. }
    \label{fig3}
\end{figure}
Through the analysis of the PTX code for Algorithm~\ref{algo1}, we identified
that branch instructions ({\em bra}) account for over 6\% of the code. This
method relies heavily on conditional branch instructions to reduce computational load,
making it unsuitable for GPUs, which are optimized for high computational
performance.
To address this issue, we adopt a method where the 16 differences $D_t$ are
pre-computed and encoded into a 32-bit unsigned integer for subsequent
classification. By performing bit-wise operations on this binary representation, we
check consistency by verifying the values, thereby replacing cumbersome
conditional classifications.
As illustrated in Fig~\ref{fig3}, the 32-bit unsigned integer, {\em
  buffer32b}, is divided into two parts to store the differences: the upper 16
bits store the dark state results, and the lower 16 bits store the bright. Each
part sequentially reflects the 16 $D_t$ values from high to low bits. When $D_t\geq{t}$, $d^i=1$
and $b^i=0$; conversely, when $D_t\leq{-t}$, $d^i=0$ and $b^i=1$. When
$-t< D_t <t$, both $d^i$ and $b^i$ are 0. This encoding strategy allows
for consistency to be determined by verifying the status of ``1'' in any
consecutive 9-bit binary number, either $d^i$ to $d^{i+9}$ or $b^i$ to
$b^{i+9}$, referred to as a {\em segment}. This alleviates the need for
conditional branch instructions, making the method more suitable for GPU execution.
Additionally, switching between different segments can be efficiently
accomplished through binary shifting, which significantly reduces instruction
execution costs compared to Algorithm~\ref{algo1}.
%

Algorithm~\ref{algo2} presents the pseudocode for the GPU kernel of the proposed
FAST feature point detection method. Initially, 16 differences $D_t$ are
compared with the threshold $t$ to generate a 32-bit buffer, {\em buffer32b}
(line 5). In the {\em bufferGeneration}, conditional operators are employed
instead of {\em if-else} statements to determine the status and repeated continuously
16 times without any other operation, enhancing operating efficiency at the
instruction level. Next, the lower 8 bits of both the dark and bright parts are
connected with their respective 16 bits to create two new 24-bit buffers: {\em
  buffer24b}$_{dark}$ and {\em buffer24b}$_{bright}$, completing the 16 segments
for each part. The purpose of these operations are still to allow the two parts
to complete the classifications for 16 segments in one loop, reducing the need for
additional branch instructions. Finally, consistency is checked
sequentially starting from the lower 9 bits of each part. If the value is {\em
  0x1FF}, which means nine continuous ``1'', indicating that the candidate pixel
$I_p$ meets the requirement of the FAST feature point. Then, the loop terminates 
and returns the classification result. Otherwise, the segment is shifted one bit to
the right and rechecked.
\begin{algorithm}
\setstretch{1.1} 
\footnotesize  
\caption{A pseudo code of the proposed FAST feature point detection}
\begin{algorithmic}[1]
\State \textbf{Input}: $I_p$, $I_i[16]$  \Comment{Candidate pixel $I_p$ and its surrouding 16 pixels $I_i$.}
\State \textbf{Output}: $isPoint$   
\State $isPoint \leftarrow False$;\Comment{Initialization.}
\State $buffer32b \leftarrow 0x0$; \Comment{32bit buffer.}
\State $buffer32b \leftarrow bufferGeneration(I_p, I_i[16])$\\ \Comment{Calculate $D_t$ and store them as Boolean values within a 32-bit buffer.}
\State $buffer24b_{dark} \leftarrow combine(buffer32b[23:16],buffer32b[31:16])$;\\  \Comment{Connect the lower 8 bits of the dark segment with itself.}
\State $buffer24b_{bright} \leftarrow combine(buffer32b[7:0],buffer32b[15:0])$;\\   \Comment{Connect the lower 8 bits of the bright segment with itself.}
\For{$k=0$ to 16}      \Comment{Check 16 segments for dark state.}
   \If{$buffer24b_{dark}\ \&\ 0x1FF==0x1FF$}\Comment{Determinate the consistency.}
     \State $isPoint \leftarrow True$;
     \State \Return $isPoint$;  \Comment{skip the remaining determinations}.
   \Else
     \State $buffer24b_{dark} >>=1$; \Comment{Shift to the next segment.}
   \EndIf
\EndFor
\For{$k=0$ to 16}  \Comment{Check 16 segments for bright state.}
   \If{$buffer24b_{bright}\ \&\ 0x1FF==0x1FF$ } \Comment{Determinate the consistency.}
     \State $isPoint \leftarrow True$;
     \State \Return $isPoint$; \Comment{skip the remaining determinations}.
   \Else
     \State $buffer24b_{bright} >>=1$; \Comment{Shift to the next segment.}
   \EndIf
\EndFor
\State \Return $isPoint$;
\end{algorithmic}
\label{algo2}
\end{algorithm}

Unlike Algorithm~\ref{algo1}, the proposed method does not pre-determine the
state $D_{flag}$; instead, it starts checking the bright state only after
completing the classification of the dark state. The primary reason for checking the
status of $D_{flag}$ in Algorithm~\ref{algo1} is to skip unnecessary classifications
and difference calculations, as the {\em Instruction Circle Time} (ICT) for
conditional branch instructions is generally longer than that for other arithmetic
instructions. In contrast, the proposed method requires fewer loops and conditional
branch instructions, ensuring high execution efficiency even when employing a
brute-force method.

\subsection{Optimization for Harris score calculation}
Based on the analysis of the results of Section~\ref{analysis:harris}, the Harris score calculation is integrated with FAST feature point detection. Specifically, when a thread detects a FAST feature point (i.e., the returned {\em isPoint} is {\em True} from Algorithm~\ref{algo2}), the Harris score is calculated immediately. This approach significantly reduces the burden of intra-frame memory accesses by avoiding repeated loading of feature points and image data.
In addition, since the Harris score calculation requires to compute pixel
gradients with the Sobel operator in a 7$\times$7 window, which is not only
highly parallelisable but also involves redundant computations of 6 columns (or
6 rows) between adjacent windows, multi-threads are driven in parallel to
accelerate the calculation for each FAST feature point. In our implementation, a
flag $SobelFlag$ is set to check whether any thread in the same warp detects a
FAST feature point as shown in the line 8 of Algorithm~\ref{algo3}. If it is
{\em True}, all threads within the same warp calculate gradients within a
7$\times$38 (32+6) window and store them in shared memory. Thus, all gradients
required by any FAST feature point can be shared across different threads without
recalculations. Although this strategy may lead to redundant invalid
calculations in extreme cases (e.g. only one feature point is detected), it will
not degrade the performance due to the warp-based GPU parallel processing
mechanism. Furthermore, since the purpose of performing a Harris corner
detection here is to eliminate redundant FAST feature points, the proposed strategy can be shown to work more effectively.
During the Harris score calculation, the {\em separable Sobel operator} is
widely used to reduce the computation complexity~\cite{chang2023multi}. The
original 2D Sobel operator is decomposed into the product of two vectors:
\begin{align}
\label{equ4}
\mathbf{G_x} &= g_x*I=
\begin{bmatrix}
-1 & 0 & 1\\
-2 & 0& 2\\
-1& 0&1\\
\end{bmatrix} * I
=\begin{bmatrix}1 \\ 2 \\ 1 \end{bmatrix}*(\begin{bmatrix} -1& 0&1 \end{bmatrix}* I)=u_x*(v_x*I)\nonumber\\
\mathbf{G_y} &= g_y*I=
\begin{bmatrix}
-1 & -2 & 1\\
0 & 0& 0\\
1& 2&1\\
\end{bmatrix} * I
=\begin{bmatrix}-1 \\ 0 \\ 1 \end{bmatrix}*(\begin{bmatrix} 1& 2& 1 \end{bmatrix}* I)=u_y*(v_y*I).\\
\nonumber
\end{align}
Thus, the image is first convolved by the decomposed 1D vectors $v_x$ and $v_y$
to calculate the gradients along the X direction, which are further convolved by
vectors $u_x$ and $u_y$ along the Y direction. 
This method is not only suitable to be executed in parallel but also
significantly reduces the calculation complexity of the Sobel operator. However,
it usually requires additional memory space to store the intermediate results
(the parts inside the brackets of Equation~\ref{equ4}). While this hardware
resource costs is typically negligible for high-end GPUs, it is very expensive
for embedded GPUs with limited on-chip resources, which can seriously impact the
thread parallelism.
\begin{figure}
    \centering \includegraphics[width=0.95\linewidth]{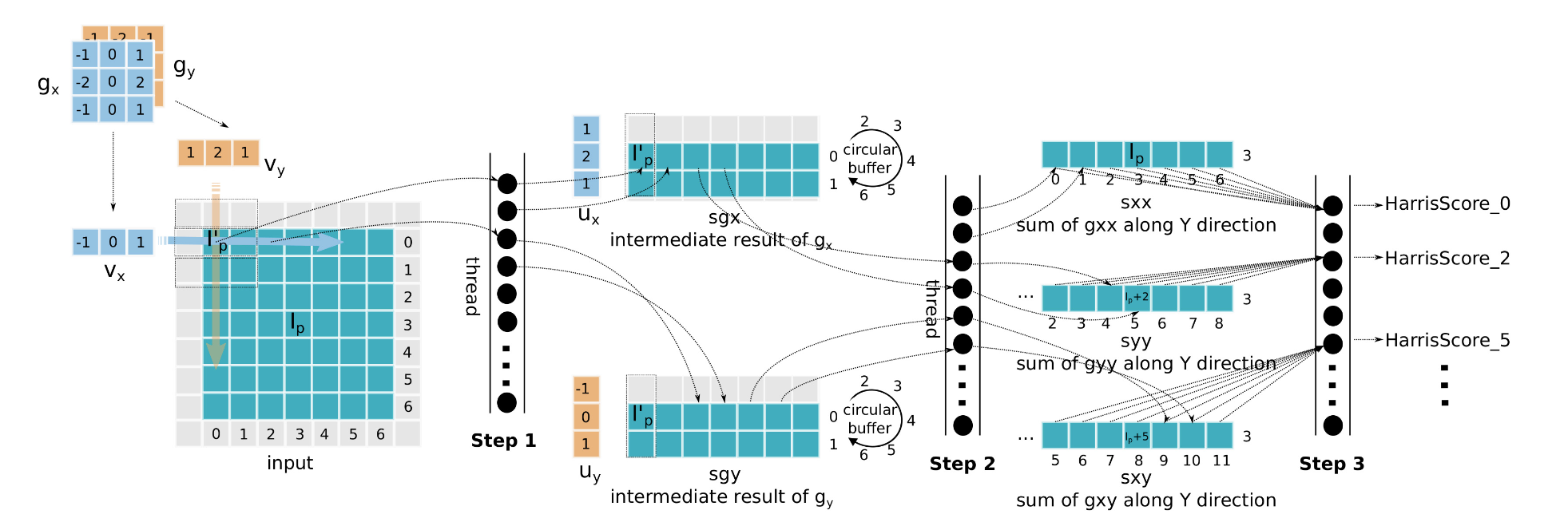}
    \caption{Processing flow of Harris score calculation. Step1: 1D Sobel
      Operator. Step2: 1D Sobel operator \& accumulation of gradients along the
      Y direction. Step3: Accumulation of gradients along the X direction.}
    \label{fig4}
\end{figure}
\begin{algorithm}
\setstretch{1.1} 
\footnotesize  
\caption{A pseudo code of Harris score calculation by semi-separable Sobel operator}
\begin{algorithmic}[1]
\State \textbf{Input}: $I$, $isPoint$  \Comment{Connect with the Algorithm~\ref{algo2}.}
\State \textbf{Output}: $HarrisScore$  \Comment{Used to determine a corner.}
\State $gx,gy,gxx,gyy,gxy$; \Comment{Gradient related variables.}
\State $sgx[3], sgy[3]$; \Comment{Circular buffers.}
\State $\_\_shared\ memory\_\_\ \ \ \ sxx[FAST\_WIDTH], syy[FAST\_WIDTH],sxy[FAST\_WIDTH]$;\\  \Comment{Story the sum of gxx,gyy and gxy along Y direciton.}
\State $SobelFlag \leftarrow \_\_any\_sync(0xFFFFFFFF, isPoint)$; \Comment{Check if any thread in a warp detects a FAST feature point.}
\If{$SobelFlag$}
  \For{$r=0$ to 2} \Comment{Step1 of Fig~\ref{fig4}: 1D Sobel Operator.}
    \State $sgx[r] \leftarrow I[r][Tid+1]-I[r][Tid-1]$;
    \State $sgy[r] \leftarrow I[r][Tid+1]+2*I[r][Tid]+I[r][Tid-1]$;
  \EndFor
  \State $gx \leftarrow sgx[0]+2*sgx[1]+sgx[2]$;
  \State $gy \leftarrow sgy[1][Tid]-sgy[0][Tid]$;
  \State $gxx=gx*gx,\ gyy=gy*gy,\ gxy=gx*gy$;\\
  \For{$r=3$ to $HarrisSize+1$}  \Comment{Step2 of Fig~\ref{fig4}: calculate and accumulate the gradients along the Y direction.}
    \State $sgx[r\%3] \leftarrow I[r][Tid+1]+2*I[r][Tid]+I[r][Tid-1]$;
    \State $sgy[r\%3] \leftarrow I[r+1][Tid]-I[r-1][Tid]$;
    \State $gx \leftarrow sgx[(r-2)\%3]+2*sgx[(r-1)\%3]+sgx[r\%3]$;
    \State $gy \leftarrow sgy[r\%3]-sgy[(r-2)\%3]$;
    \State $gxx+=gx*gx,\ gyy+=gy*gy,\ gxy+=gx*gy$;
  \EndFor
  \State $sxx[Tid] \leftarrow gxx,\ syy[Tid] \leftarrow gyy,\ sxy[Tid] \leftarrow gxy$;
\EndIf
\State
\State $gxx, gyy, gxy \leftarrow 0$; 
\If{$isPoint$} \Comment{Only calculate Harris scores for Fast points.}
  \For{$j=0$ to $HarrisSize$}  \Comment{Step3 of Fig~\ref{fig4}: accumulate the gradients along the X direction.}
     \State $gxx+=sxx[Tid+j],\ gyy+=syy[Tid+j],\ gxy+=sxy[Tid+j]$;
  \EndFor
  \State $HarrisScore \leftarrow (gxx*gyy-gxy*gxy-k*(gxx+gyy)*(gxx+gyy)*factor$;\\ \Comment{Harris score calculation according to Equation~\ref{equ3}.}
\EndIf

\State \Return $HarrisScore$;
\end{algorithmic}
\label{algo3}
\end{algorithm}

To address this issue, a {\em semi-separable Sobel operator} is proposed that
keeping the decomposition mechanism without storing all intermediate results in
the X direction. Unlike the original method, where the convolution along the X
direction is completely executed before the Y direction, the proposed method alternately
executes the convolutions in both directions. As shown in Fig~\ref{fig4}, the
top three rows are first convolved by 1D vectors $v_1$ and $v_2$, and save the
results in separate registers (Step 1). As a {\em circular buffer}, these
registers are driven by each thread to support the convolution of the remaining rows in a sliding manner. When a new convolution is completed by the vector
$v$, the oldest result in the circular buffer is updated to complete the
convolution by the vector $u$. At the same time, gradients in the Y direction
can be accumulated together during the sliding process (Step 2). Finally, each
thread horizontally accumulates obtained gradients within the same window to
calculate a Harris score.
Algorithm~\ref{algo3} illustrates the processing details of the proposed semi-separable Sobel operator. It is executed directly following
Algorithm~\ref{algo2} when {\em isPoint} is {\em True}. To optimize the memory
usage, each thread employs six registers, $sgx[3]$ and $sgy[3]$ (line 4), as
circular buffers to store intermediate results in both directions. Since threads
are executed along the X direction, three shared memory arrays, $sxx$, $syy$,
$sxy$, are allocated to accumulate the gradients in the Y direction (line
5). Upon detecting a FAST feature point, the top three rows are first convolved
and the results are stored in their respective registers (line 9 to 12). Then,
after completing the convolution in the Y direction, parts of the $gxx$, $gyy$
and $gxy$ items required in Equation~\ref{equ3} are calculated (line 13 to
15). The remaining parts are calculated in a sliding manner, with accumulating
results of each row after each update (line 17 to 23). Here, the incremental
rows are indexed by $r\%3$. 
Finally, the thread that detected a FAST feature point accumulates the intermediate
results stored in the shared memory along the X direction and calculates the
final Harris score according to Equation~\ref{equ3} (line 27 to 32).

\section{Experimental evaluation}
\label{sec:experiment}
\begin{figure}
    \centering \includegraphics[width=0.9\linewidth]{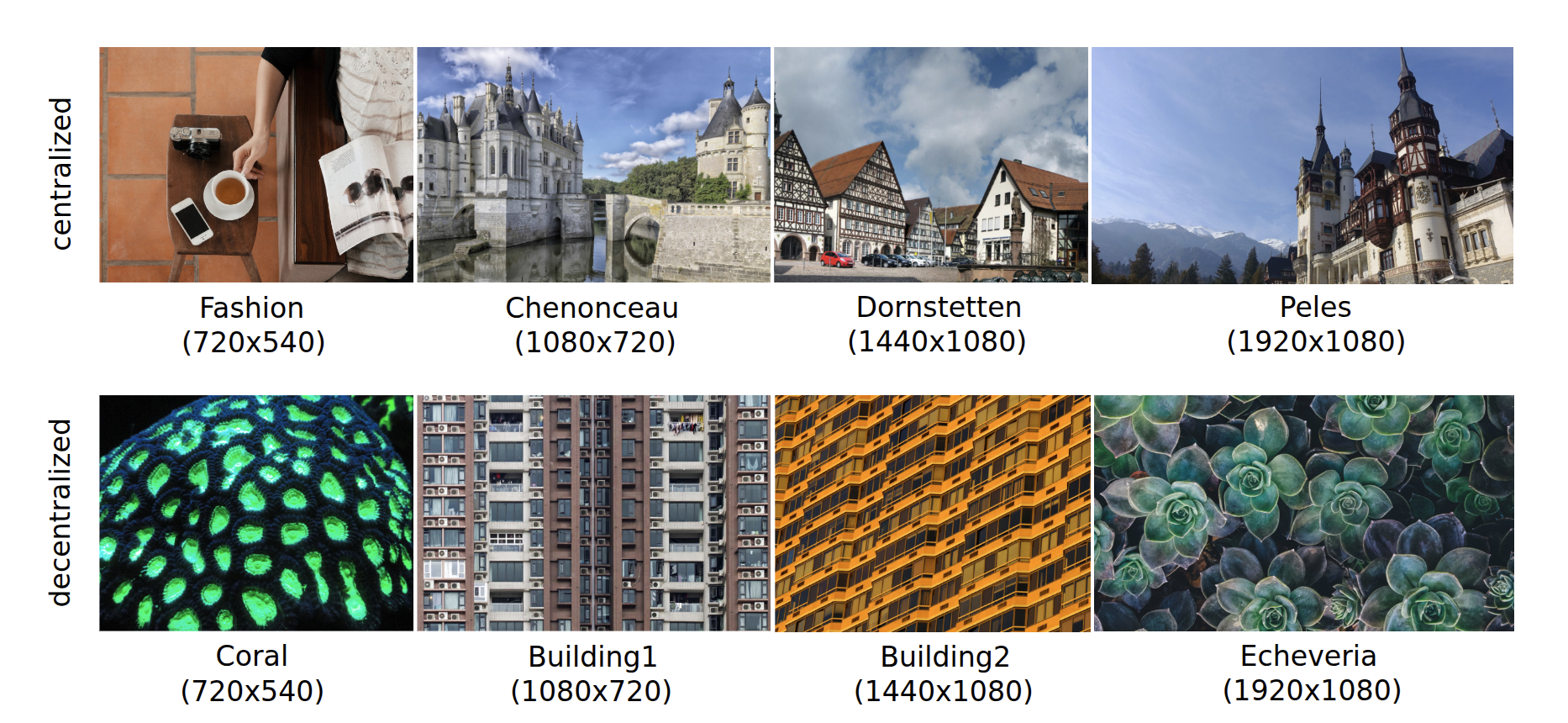}
    \caption{Image set used in the experiments}
    \label{fig5}
\end{figure}

We thoroughly evaluated the proposed acceleration methods on a Jetson TX2 GPU and analyzed the impact of each step on performance
improvement. As shown in Table~\ref{tb1}, the Jetson TX2 GPU features the NVIDIA
Pascal Architecture with 256 CUDA cores, supporting by a quad-core ARM
Cortex-A57 MPCore, ensuring efficient and powerful computing performance. The
system is equipped with 8GB of 128-bit LPDDR4 memory operating at 1866 MHz,
providing a bandwidth of 59.7 GB/s. Compared to the Jetson Nano and Xavier, the
Jetson TX2 usually provides a significant upgrade over the Nano while being more
power-efficient and cost-effective than the Xavier. The good balance makes the
TX2 suitable for a wide range of applications, such as robotic vision and
wearable devices, offering sufficient power for complex AI tasks while
maintaining efficiency.
The proposed strategy is implemented with CUDA~\cite{cuda} and runs on Ubuntu 18.04 LTS, complied with NVCC 10.2 using the "-arch=sm\_62" flag.
Figure~\ref{fig5} shows eight selected images to evaluate the performance of
the optimization methods. These images have different resolutions, including 720$\times$540, 
1080$\times$720, 1440$\times$1080 and
1920$\times$1080, and different numbers of FAST feature points from 4900 to
77000 (detected by OpencvCPU library with four-layer pyramid). According to the distribution 
of FAST feature points, these images can be divided into two categories: {\em centralized} and {\em
  decentralized}. Due to the
Harris corner detection directly following the FAST
detection, the feature points distribution is crucial to the parallelism of GPU
threads. 
\begin{table*}
\caption{Jetson TX2 specifications \& system environment}
\footnotesize
\begin{tabular}{c|c|c|c} 
\toprule
\textbf{GPU}& NVIDIA Pascal$^{TM}$ GPU architecture &\textbf{Storage}& 32GB eMMC 5.1\\
\cmidrule(lr){1-4}
\textbf{Cuda Cores}&256&\textbf{Power}& 7.5W/15W\\
\cmidrule(lr){1-4}
\textbf{Boost Clock}& 1.3GHz&\textbf{OS}& Ubuntu 18.04 LTS\\	
\cmidrule(lr){1-4}
\multirow{2}{*}{\bf CPU}&Dual-Core NVIDIA Denver 2 64-Bit &\multirow{2}{*}{\bf CUDA}& \multirow{2}{*}{10.2}\\
& Quad-Core ARM Cortex-A57 MPCore&&\\
\cmidrule(lr){1-4}
\multirow{2}{*}{\bf Memory}& 8GB 128-bit LPDDR4 Memory &\multirow{2}{*}{\bf OpenCV}& \multirow{2}{*}{4.5.3}\\
& 1866 MHx-59.7 GB/s &&\\
\bottomrule
\end{tabular}
\label{tb1}
\end{table*}
\subsection{Evaluation of Proposed Oriented FAST Kernel}
To clarify the acceleration effect brought by each optimization step, the proposed methods are categorized into the several components and choose the standard implementation CUDA\_ORB~\cite{cuda-ORB} as the baseline to perform extensive experiments.

As shown in Table~\ref{tb2}, these methods are divided into two classes: the FAST-like methods and the Harris-like methods.
 Within each class, the methods are then compared with the
baseline. Since the FAST detector in CUDA\_ORB completely relies on global
memory, these FAST-like methods are defined as following: {\em BinaryFAST},
which implements the proposed binary method on global memory; {\em SMFAST},
which involves moving the FAST detector to shared memory; and the {\em
  SM+BinaryFAST}, which implements the binary method on shared memory.
For the Harris-like methods, they are still defined as following:
{\em SepHarris}, where each thread independently completes the separable Sobel
operation in a 2D square window; {\em Para-SepHarris}, which performs the
separable Sobel operation in parallel, storing all intermediate results on
shared memory; {\em Semi-SepHarris}, which implements the separable Sobel
operation in parallel but saves memory usage using a circular buffer.

Table~\ref{tb3} lists the counts of primary instructions in the PTX codes for
various methods. Here, only the FAST and Harris function codes are shown
here and counts exclude loop unrolling. These instructions include
{\em load} and {\em store} for both on-chip and off-chip memories,
the primary arithmetic instructions {\em add} and {\em sub} utilized in Harris
detection, and the most time-consuming branch instruction {\em bra}.
In Table~\ref{tb3}, the methods {\em Baseline(FAST)}, {\em BinaryFAST} and {\em
  Baseline(Harris)} execute higher numbers of {\em ld.global} instructions than
other methods because all of them completely rely on global memory. Conversely,
other methods effectively utilize shared memory, resulting in a greater number
of {\em ld.shared} instructions. The utilization of {\em ld.const} in methods {\em
  Baseline(FAST)}, {\em BinaryFAST} and {\em SMFAST} is mainly for the flexible
calculation of pixel address indexes. Although the constant memory can be
accessed efficiently, statics constants are defined directly in the
implementation because of the relatively small number of constants and cache hit
rate considerations. Moreover, compared with the {\em Baseline(FAST)} and {\em
  SMFAST}, the proposed methods {\em BinaryFAST} and {\em SM-BinaryFAST} significantly
reduce the use of the branch instruction {\em bra}. Here, it is noteworthy that
the {\em Para-SepHarris} employs more {\em st.shared} operations than other
methods because it stores all intermediate results from the Sobel
operations. Finally, the total count of the {\em Baseline(Harris)} is
considerably smaller than that of other methods because it nests the
two-dimensional Sobel operations within two loops without unrolling them.
\begin{table*}
\caption{Definition of different methods for performance comparison.}
\footnotesize
\begin{tabular}{r|r|l} 
\toprule
&\textbf{Baseline}&The original CUDA\_ORB method implemented by~\cite{cuda-ORB}.  \\
\cmidrule(lr){1-3}
\multirow{3}{*}&\textbf{BinaryFAST}&FAST detector based on the binary encoding strategy.  \\
\cmidrule(lr){2-3}
{\bf FAST}&\textbf{SMFAST}&FAST detector implemented on shared memory.  \\
\cmidrule(lr){2-3}
&\textbf{SM-BinaryFAST}&{\em BinaryFAST} implemented on shared memory.   \\
\cmidrule(lr){1-3}
\multirow{3}{*}&\textbf{SepHarris}&Harris detector based on separable Sobel operator.  \\
\cmidrule(lr){2-3}
{\bf Harris}&\textbf{Para-SepHarris}&Semi-separable Sobel operator that stores
  all intermediate results on-chip. \\
\cmidrule(lr){2-3}
&\textbf{Semi-SepHarris}&Semi-separable Sobel operator using a
  circular buffer.  \\
\bottomrule
\end{tabular}
\label{tb2}
\end{table*}

\begin{table*}
\footnotesize
\caption{Comparison of the number of primary instructions in PTX code} \scriptsize
\begin{tabular}{r|c|c|c|c|c|c|c|c|c} 
\toprule
&{\bf \#ld.global}&{\bf \#ld.const}&{\bf \#ld.shared}&{\bf \#st.global}&{\bf \#st.shared}&{\bf \#add}&{\bf \#sub}&{\bf \#bra}&{\bf \#Total}\\
\cmidrule(lr){1-10}
{\bf Baseline(FAST)}&27&42$^1$&0&2&0&-&-&29$^2$&422\\
\cmidrule(lr){1-10}
{\bf BinaryFAST}&17&16&0&2&0&-&-&19$^3$&366\\
\cmidrule(lr){1-10}
{\bf SMFAST}&9&42$^1$&27&2&9&-&-&37$^2$&512\\
\cmidrule(lr){1-10}
{\bf SM-BinaryFAST}&13&0&17&2&18&-&-&15$^3$&393\\
\cmidrule(lr){1-10}
{\bf Baseline(Harris)}&9&-&0&2&0&21&8&-&111$^4$\\
\cmidrule(lr){1-10}
{\bf SepHarris}&1&0&27&2&0&41&17&-&213\\
\cmidrule(lr){1-10}
{\bf Para-SepHarris}&1&0&85&2&42$^5$&83&21&-&429\\
\cmidrule(lr){1-10}
{\bf Semi-SepHarris}&1&0&75&2&6$^6$&91&33&-&380\\
\bottomrule
\end{tabular}
\begin{tablenotes}
\footnotesize
\item \#: counts of code line; $^1$: load constant for address indexing; $^2$: more branch instructions; $^3$: fewer branch instructions; 
\item $^4$: repeating operations in loops; $^5$: save intermediate results; $^6$: usage of circular buffer;
\end{tablenotes}
\label{tb3}
\end{table*} %

Due to the basic FAST detection in CUDA\_ORB requires a prejudgment regarding the
shape of pattern, we designed five test cases to ensure a comprehensive
evaluation, as shown in Fig~\ref{fig6}. These test cases include three
different continuous patterns (cases 1, 2 and 3) and two discontinuous patterns
(cases 4, 5) to adapt to different conditions.
For each pattern, we generated a dedicated test image composed of 625 identical
patterns within a 1024$\times$1024 image. This aims to clarify the processing
performance of the same method across different patterns, while maintaining
the parallelism. Figure~\ref{fig7} shows the evaluation results under different
test cases. The image size of each layer
decreases exponentially to accommodate different feature sizes. In Fig~\ref{fig7}, the runtime of the four-layer pyramid is
slower than that of the single-layer, but not by more than twice. In all
cases, the proposed binary-based methods demonstrate stable efficiency because of their
executions independence of the pattern shape. On the other hand, the runtime of
the {\em Baseline(FAST)} in case 3 is longer than that of other cases because it
requires an additional half-circle judgment. In cases 4 and 5, the prejudgment
mechanism of the {\em Baseline(FAST)} plays a role, allowing direct skipping of
the unnecessary processing to save time. Compared to the baseline methods {\em
  Baseline(FAST)} and {\em SMFAST}, the {\em SM+BinaryFAST} shows better
performance, whereas the {\em BinaryFAST} performs less favorably. This
discrepancy arises because the proposed method is designed to minimize the use of branch
statements rather than reducing data access. Therefore, frequent global memory
access results in bandwidth latency, negatively impacting performance.
\begin{figure}
  \centering \includegraphics[width=0.9\linewidth]{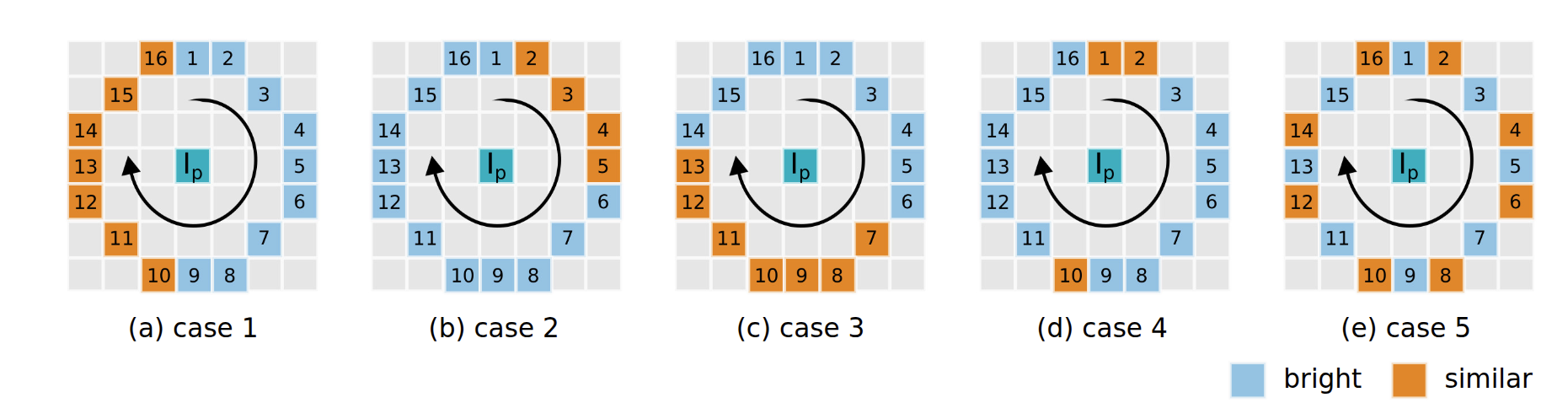}
  \caption{Test cases for FAST detection. (a) nine continuous bright points from
    the beginning; (b) more than nine continuous bright points from 6 to 1; (c) nine
    continuous bright points over the 16th pixel; (d) many continuous bright
    points but less than nine; (e) no continuous bright point.}
  \label{fig6}
\end{figure}
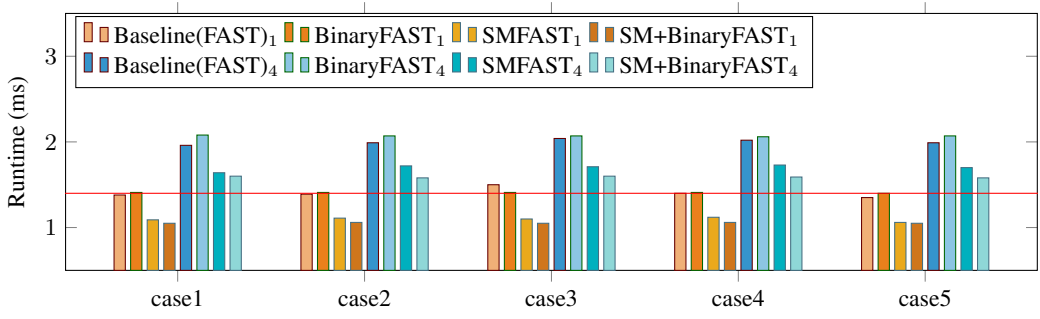
\begin{figure}
\centering
\footnotesize
\begin{tikzpicture}[scale=1]
\begin{axis}[
    bar width=0.15cm,
    ybar,
    width=14.5cm,
    height=5cm,
    ymin=0.5,
    ymax=3.5,
    ylabel={Runtime (ms)}, 
    symbolic x coords={case1, case2, case3, case4, case5},
    xtick={case1, case2, case3, case4, case5},
    xtick distance = 1cm,
    enlarge x limits=0.15,
    legend style={at={(0.01,0.99)},anchor=north west,legend columns=4}
    ]

\addplot [black!60!red,fill=color4] coordinates  { (case1,	1.38
) (case2,	1.39
) (case3,	1.50
) (case4,	1.40
) (case5,	1.35
)}; \addlegendentry{Baseline(FAST)$_{1}$} 

\addplot [black!60!green ,fill=color2] coordinates {(case1,	1.41
) (case2,	1.41
) (case3,	1.41
) (case4,	1.41
) (case5,	1.40
)}; \addlegendentry{BinaryFAST$_{1}$} 

\addplot [black!60!cyan, fill=color7] coordinates {(case1,	1.09
) (case2,	1.11
) (case3,	1.10
) (case4,	1.12
) (case5,	1.06
)}; \addlegendentry{SMFAST$_{1}$}	

\addplot [black!60!cyan, fill=color8] coordinates {(case1,	1.05
) (case2,	1.06
) (case3,	1.05
) (case4,	1.06
) (case5,	1.05
)}; \addlegendentry{SM+BinaryFAST$_{1}$}

\addplot [black!60!red,fill=color1] coordinates  { (case1,	1.96
) (case2,	1.99
) (case3,	2.04
) (case4,	2.02
) (case5,	1.99
)}; \addlegendentry{Baseline(FAST)$_{4}$} 

\addplot [black!60!green ,fill=color3] coordinates {(case1,	2.08
) (case2,	2.07
) (case3,	2.07
) (case4,	2.06
) (case5,	2.07
)}; \addlegendentry{BinaryFAST$_{4}$} 

\addplot [black!60!cyan, fill=color5] coordinates {(case1,	1.64
) (case2,	1.72
) (case3,	1.71
) (case4,	1.73
) (case5,	1.70
)}; \addlegendentry{SMFAST$_{4}$}	

\addplot [black!60!cyan, fill=color6] coordinates {(case1,	1.60
) (case2,	1.58
) (case3,	1.60
) (case4,	1.59
) (case5,	1.58
)}; \addlegendentry{SM+BinaryFAST$_{4}$}	
\draw[red] (-120,90) -- (520,90);
\end{axis}
\end{tikzpicture}
\caption{Performance comparison of FAST-like algorithms on different test cases. The subscripts 1 and 4 represent the number of layers in the pyramid constructed during the FAST point detection.}
\label{fig7}
\end{figure}
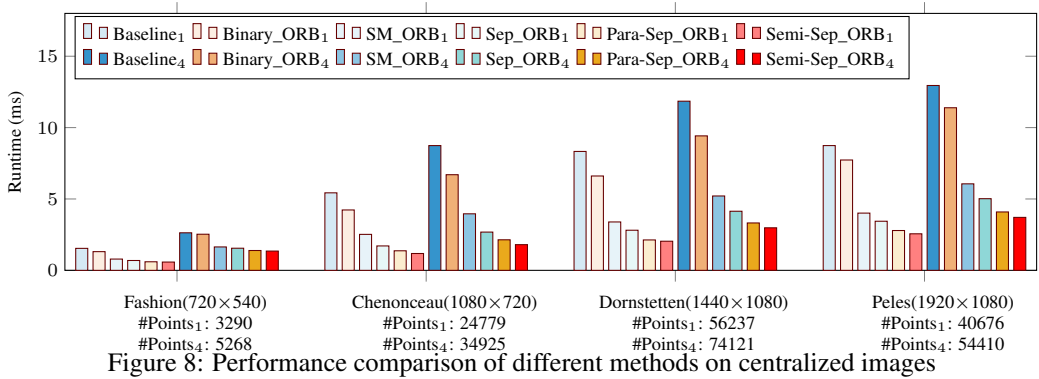
\begin{figure}
\centering
\scriptsize
\begin{tikzpicture}[scale=1]
\begin{axis}[
    bar width=0.16cm,
    ybar,
    width=14.5cm,
    height=5cm,
    ymin=0,
    ymax=18,
    ylabel={Runtime (ms)}, 
    symbolic x coords={Fashion(720$\times$540), Chenonceau(1080$\times$720),Dornstetten(1440$\times$1080),Peles(1920$\times$1080)},
    xtick={Fashion(720$\times$540), Chenonceau(1080$\times$720),Dornstetten(1440$\times$1080),Peles(1920$\times$1080)},
    xticklabels={{\node[align=center]{Fashion(720$\times$540)\\\#Points$_1$: 3290\\\#Points$_4$: 5268};}, {\node[align=center]{Chenonceau(1080$\times$720)\\\#Points$_1$: 24779\\\#Points$_4$: 34925};}, {\node[align=center]{Dornstetten(1440$\times$1080)\\\#Points$_1$: 56237\\\#Points$_4$: 74121};}, {\node[align=center]{Peles(1920$\times$1080)\\\#Points$_1$: 40676\\\#Points$_4$: 54410};}},
    xtick distance = 1cm,
    enlarge x limits=0.15,
    legend style={at={(0.01,0.99)},anchor=north west,legend columns=6}
    ]

\addplot [black!60!red,fill=color1!70!white!30] coordinates  { (Fashion(720$\times$540), 1.54
) (Chenonceau(1080$\times$720),	5.43
) (Dornstetten(1440$\times$1080),	8.33
) (Peles(1920$\times$1080),	8.74
)}; \addlegendentry{Baseline$_{1}$}

\addplot [black!60!red,fill=color4!70!white!30] coordinates  { (Fashion(720$\times$540),	1.31
) (Chenonceau(1080$\times$720),	4.23
) (Dornstetten(1440$\times$1080),	6.61
) (Peles(1920$\times$1080),	7.73
)}; \addlegendentry{Binary\_ORB$_{1}$}

\addplot [black!60!red,fill=color3!70!white!30] coordinates  { (Fashion(720$\times$540),	0.79
) (Chenonceau(1080$\times$720),	2.52
) (Dornstetten(1440$\times$1080),	3.39
) (Peles(1920$\times$1080),	4.01
)}; \addlegendentry{SM\_ORB$_{1}$} 

\addplot [black!60!red,fill=color6!70!white!30] coordinates  { (Fashion(720$\times$540),	0.69
) (Chenonceau(1080$\times$720),	1.71
) (Dornstetten(1440$\times$1080),	2.81
) (Peles(1920$\times$1080),	3.44
)}; \addlegendentry{Sep\_ORB$_{1}$} 

\addplot [black!60!red,fill=color7!70!white!30] coordinates  { (Fashion(720$\times$540),	0.6
) (Chenonceau(1080$\times$720),	1.37
) (Dornstetten(1440$\times$1080),	2.13
) (Peles(1920$\times$1080),	2.79
)}; \addlegendentry{Para-Sep\_ORB$_{1}$} 

\addplot [black!60!red,fill=red!50!white] coordinates  { (Fashion(720$\times$540),	0.58
) (Chenonceau(1080$\times$720),	1.18
) (Dornstetten(1440$\times$1080),	2.04
) (Peles(1920$\times$1080),	2.56
)}; \addlegendentry{Semi-Sep\_ORB$_{1}$}

\addplot [black!60!red,fill=color1] coordinates  { (Fashion(720$\times$540),	2.63
) (Chenonceau(1080$\times$720),     8.74
) (Dornstetten(1440$\times$1080),	11.85
) (Peles(1920$\times$1080),	12.95
)}; \addlegendentry{Baseline$_{4}$} 

\addplot [black!60!red,fill=color4] coordinates  { (Fashion(720$\times$540),	2.53
) (Chenonceau(1080$\times$720),	6.7
) (Dornstetten(1440$\times$1080),	9.42
) (Peles(1920$\times$1080),	11.39
)}; \addlegendentry{Binary\_ORB$_{4}$} 

\addplot [black!60!red,fill=color3] coordinates  { (Fashion(720$\times$540),	1.64
) (Chenonceau(1080$\times$720),	3.96
) (Dornstetten(1440$\times$1080),    5.21
) (Peles(1920$\times$1080),	6.06
)}; \addlegendentry{SM\_ORB$_{4}$} 

\addplot [black!60!red,fill=color6] coordinates  { (Fashion(720$\times$540),	1.55
) (Chenonceau(1080$\times$720),	2.68
) (Dornstetten(1440$\times$1080),	4.14
) (Peles(1920$\times$1080),	5.02
)}; \addlegendentry{Sep\_ORB$_{4}$} 

\addplot [black!60!red,fill=color7] coordinates  { (Fashion(720$\times$540),	1.39
) (Chenonceau(1080$\times$720),	2.14
) (Dornstetten(1440$\times$1080),	3.32
) (Peles(1920$\times$1080),	4.09
)}; \addlegendentry{Para-Sep\_ORB$_{4}$} 

\addplot [black!60!red,fill=red] coordinates  { (Fashion(720$\times$540),	1.35
) (Chenonceau(1080$\times$720),	1.8
) (Dornstetten(1440$\times$1080),	2.98
) (Peles(1920$\times$1080),	3.71
)}; \addlegendentry{Semi-Sep\_ORB$_{4}$} 

\end{axis}
\end{tikzpicture}
\vspace*{0.5cm}
\caption{Performance comparison of different methods on centralized images}
\label{fig8}
\end{figure}

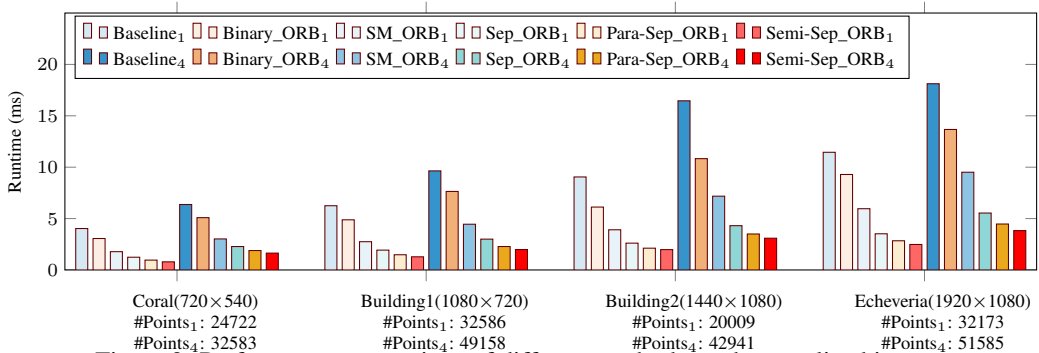
\begin{figure}
\centering
\scriptsize
\begin{tikzpicture}[scale=1]
\begin{axis}[
    bar width=0.16cm,
    ybar,
    width=14.5cm,
    height=5cm,
    ymin=0,
    ymax=25,
    ylabel={Runtime (ms)}, 
    ytick={0,5,10,15,20},
    symbolic x coords={Coral(720$\times$540), Building1(1080$\times$720),Building2(1440$\times$1080),Echeveria(1920$\times$1080)},
    xtick={Coral(720$\times$540), Building1(1080$\times$720),Building2(1440$\times$1080),Echeveria(1920$\times$1080)},
    xticklabels={{\node[align=center]{Coral(720$\times$540)\\\#Points$_1$: 24722\\\#Points$_4$: 32583};}, {\node[align=center]{Building1(1080$\times$720)\\\#Points$_1$: 32586\\\#Points$_4$: 49158};},{\node[align=center]{Building2(1440$\times$1080)\\\#Points$_1$: 20009\\\#Points$_4$: 42941};},{\node[align=center]{Echeveria(1920$\times$1080)\\\#Points$_1$: 32173\\\#Points$_4$: 51585};}},
    xtick distance = 1cm,
    enlarge x limits=0.15,
    legend style={at={(0.01,0.99)},anchor=north west,legend columns=6}
    ]

\addplot [black!60!red,fill=color1!70!white!30] coordinates  { (Coral(720$\times$540),	4.03
) (Building1(1080$\times$720),	6.25
) (Building2(1440$\times$1080),	9.05
) (Echeveria(1920$\times$1080),	11.45
)}; \addlegendentry{Baseline$_{1}$} 

\addplot [black!60!red,fill=color4!70!white!30] coordinates  { (Coral(720$\times$540), 3.06
) (Building1(1080$\times$720),	4.88
) (Building2(1440$\times$1080),	6.12
) (Echeveria(1920$\times$1080),	9.29
)}; \addlegendentry{Binary\_ORB$_{1}$} 

\addplot [black!60!red,fill=color3!70!white!30] coordinates  { (Coral(720$\times$540), 1.78
) (Building1(1080$\times$720),	2.75
) (Building2(1440$\times$1080),	3.91
) (Echeveria(1920$\times$1080),	5.96
)}; \addlegendentry{SM\_ORB$_{1}$} 

\addplot [black!60!red,fill=color6!70!white!30] coordinates  { (Coral(720$\times$540),	1.24
) (Building1(1080$\times$720),	1.93
) (Building2(1440$\times$1080),	2.61
) (Echeveria(1920$\times$1080),	3.52
)}; \addlegendentry{Sep\_ORB$_{1}$} 

\addplot [black!60!red,fill=color7!70!white!30] coordinates  { (Coral(720$\times$540), 0.96
) (Building1(1080$\times$720),	1.48
) (Building2(1440$\times$1080),	2.12
) (Echeveria(1920$\times$1080),	2.84
)}; \addlegendentry{Para-Sep\_ORB$_{1}$} 

\addplot [black!60!red,fill=red!60!white] coordinates  { (Coral(720$\times$540), 0.79
) (Building1(1080$\times$720),	1.28
) (Building2(1440$\times$1080),	1.98
) (Echeveria(1920$\times$1080),	2.48
)}; \addlegendentry{Semi-Sep\_ORB$_{1}$}

\addplot [black!60!red,fill=color1] coordinates  { (Coral(720$\times$540),	6.37
) (Building1(1080$\times$720),	9.64
) (Building2(1440$\times$1080),	16.46
) (Echeveria(1920$\times$1080),	18.12
)}; \addlegendentry{Baseline$_{4}$} 

\addplot [black!60!red,fill=color4] coordinates  { (Coral(720$\times$540),     5.09
) (Building1(1080$\times$720),	7.64
) (Building2(1440$\times$1080),	10.83
) (Echeveria(1920$\times$1080),	13.67
)}; \addlegendentry{Binary\_ORB$_{4}$} 

\addplot [black!60!red,fill=color3] coordinates  { (Coral(720$\times$540),	3.02
) (Building1(1080$\times$720),	4.45
) (Building2(1440$\times$1080),	7.18
) (Echeveria(1920$\times$1080),	9.51
)}; \addlegendentry{SM\_ORB$_{4}$}

\addplot [black!60!red,fill=color6] coordinates  { (Coral(720$\times$540),	2.28
) (Building1(1080$\times$720),	3.00
) (Building2(1440$\times$1080),	4.31
) (Echeveria(1920$\times$1080),	5.54
)}; \addlegendentry{Sep\_ORB$_{4}$} 

\addplot [black!60!red,fill=color7] coordinates  { (Coral(720$\times$540),	1.89
) (Building1(1080$\times$720),	2.28
) (Building2(1440$\times$1080),	3.50
) (Echeveria(1920$\times$1080),	4.47
)}; \addlegendentry{Para-Sep\_ORB$_{4}$} 

\addplot [black!60!red,fill=red] coordinates  { (Coral(720$\times$540),	1.64
) (Building1(1080$\times$720),	1.99
) (Building2(1440$\times$1080),	3.09
) (Echeveria(1920$\times$1080),   3.83
)}; \addlegendentry{Semi-Sep\_ORB$_{4}$}

\end{axis}
\end{tikzpicture}
\vspace*{0.5cm}
\caption{Performance comparison of different methods on decentralized images}
\label{fig9}
\end{figure}

To evaluate the performance of the Harris detection, we combined it with the FAST detection as the input. 
For different FAST-like methods, {\em Baseline},
{\em Binary\_ORB}, and {\em SM\_ORB} employ the same original 2D Sobel operator
to calculate the Harris score. Conversely, for different Harris-like methods, {\em
  Sep\_ORB}, {\em Para-Sep\_ORB}, and {\em Semi-Sep\_ORB} employ the same {\em
  SM-BinaryFAST} method to detect FAST feature points. Figures~\ref{fig8}
and~\ref{fig9} illustrate the runtime for eight different images shown in
Fig~\ref{fig5}. {\em \#Points} represents the FAST point number detected by
{\em Semi-Sep\_ORB}. It can be observed that regardless of the image, the
performance follows the order of {\em Baseline}<{\em Binary\_ORB}<{\em
  SM\_ORB}<{\em Sep\_ORB}<{\em Para-Sep\_ORB}<{\em Semi-Sep\_ORB}, with {\em
  SM\_ORB}, {\em Sep\_ORB}, {\em Para-Sep\_ORB} and {\em Semi-Sep\_ORB} all
using the shared memory. 
In all cases, as the image size increases, the runtime of each method also
increases. The runtime of the
FAST detection is related to the image size, while the Harris detection depends
on the number of FAST feature points. 
The performances of the {\em Baseline} methods fluctuate significantly across different images, largely due to the FAST point number.
In contrast, the proposed {\em Semi-Sep\_ORB} methods achieve consistently lower and more stable runtimes. 
These finding indicate that Harris
detection is a major time-consuming factor in Oriented FAST computations, while also demonstrating the robustness and effectiveness of the proposed optimization approach.

\begin{table*}
\captionsetup{justification=centering}
\caption{Step-by-step performance improvement percentages} \footnotesize
\begin{tabular}{r|c|c|c|c|c|c} 
\toprule
&{\bf Pyramid}&{\bf Binary\_ORB}&{\bf SM\_ORB}&{\bf Sep\_ORB}&{\bf Para-Sep\_ORB}&{\bf Semi-Sep\_ORB}\\
\cmidrule(lr){1-7}
\multirow{2}{*}{\bf Fashion}&1&+14.93\%&+39.69\%&+12.65\%&+13.04\%&+3.33\%\\
&4&+3.80\%&+35.17\%&+5.48\%&+10.32\%&+2.87\%\\
\cmidrule(lr){1-7}
\multirow{2}{*}{\bf Chenonceau}&1&+22.09\%&+40.42\%&+32.14\%&+19.88\%&+13.86\%\\
&4&+23.34\%&+40.89\%&+32.32\%&+20.14\%&+15.88\%\\
\cmidrule(lr){1-7}
\multirow{2}{*}{\bf Dornstetten}&1&+20.64\%&+48.71\%&+17.10\%&+24.19\%&+4.22\%\\
&4&+20.50\%&+44.69\%&+20.53\%&+19.80\%&+10.24\%\\
\cmidrule(lr){1-7}
\multirow{2}{*}{\bf Peles}&1&+11.55\%&+48.12\%&+14.21\%&+18.89\%&+8.24\%\\
&4&+12.04\%&+46.79\%&+17.16\%&+18.52\%&+9.29\%\\
\cmidrule(lr){1-7}
\multirow{2}{*}{\bf Coral}&1&+24.06\%&+41.83\%&+30.33\%&+22.58\%&+17.70\%\\
&4&+20.09\%&+40.66\%&+24.50\%&+17.10\%&+13.22\%\\
\cmidrule(lr){1-7}
\multirow{2}{*}{\bf Building1}&1&+21.92\%&+43.64\%&+29.81\%&+23.31\%&+13.51\%\\
&4&+20.74\%&+41.75\%&+32.58\%&+24\%&+12.71\%\\
\cmidrule(lr){1-7}
\multirow{2}{*}{\bf Building2}&1&+32.37\%&+36.11\%&+33.24\%&+18.77\%&+6.60\%\\
&4&+34.20\%&+33.70\%&+39.97\%&+18.79\%&+11.71\%\\
\cmidrule(lr){1-7}
\multirow{2}{*}{\bf Echeveria}&1&+18.86\%&+29.09\%&+21.28\%&+5.90\%&+3.15\%\\
&4&+24.54\%&+22.96\%&+21.91\%&+5.89\%&+3.52\%\\
\bottomrule
\end{tabular}
\begin{tablenotes}
\footnotesize
\item For {\em Binary\_ORB}, the percentage improvements are calculated relative to the {\em Baseline} methods shown in Figures~\ref{fig8} and \ref{fig9}. 
\item For all other methods, their improvement percentages are calculated with respect to the methods described in their previous columns.
\end{tablenotes}
\label{tb4}
\end{table*} %

Table~\ref{tb4} illustrates the performance improvement percentage of each
method shown in Figs~\ref{fig8} and~\ref{fig9}. The improvement percentage of
the same method varies little across different images, demonstrating the strong
robustness of the proposed optimization strategies. All methods effectively improved the
processing speed of Oriented FAST detector, from 2.87\% to 48.71\%. Among them,
the use of shared memory plays a crucial role in this process, improving performance by nearly 50\%. Furthermore, the proposed {\em Semi-Sep\_ORB} also achieved
up to a 17.70\% performance improvement based on the {\em Para-Sep\_ORB}.
%
\begin{figure}
\centering
\scriptsize
\begin{tikzpicture}[scale=1]
\begin{axis}[
    bar width=0.3cm,
    ybar,
    width=14.5cm,
    height=5.5cm,
    ymin=0,
    ymax=35,
    ylabel={Runtime (ms)}, 
    symbolic x coords={Fashion(720$\times$540), Chenonceau(1080$\times$720),Dornstetten(1440$\times$1080),Peles(1920$\times$1080)},
    xtick={Fashion(720$\times$540), Chenonceau(1080$\times$720),Dornstetten(1440$\times$1080),Peles(1920$\times$1080)},
    xticklabels={{\node[align=center]{Fashion(720$\times$540)\\\#Points$_1$: 3290\\\#Points$_4$: 5268};}, {\node[align=center]{Chenonceau(1080$\times$720)\\\#Points$_1$: 24779\\\#Points$_4$: 34925};}, {\node[align=center]{Dornstetten(1440$\times$1080)\\\#Points$_1$: 56237\\\#Points$_4$: 74121};}, {\node[align=center]{Peles(1920$\times$1080)\\\#Points$_1$: 40676\\\#Points$_4$: 54410};}},
    xtick distance = 1cm,
    enlarge x limits=0.15,
    legend style={at={(0.01,0.99)},anchor=north west,legend columns=4}
    ]

\addplot [black!60!red,fill=color1!70!white!30] coordinates  { (Fashion(720$\times$540), 3.07
) (Chenonceau(1080$\times$720),	10.13
) (Dornstetten(1440$\times$1080),	16.95
) (Peles(1920$\times$1080),	17.48
)};
\addlegendentry{OPENCVCPU\_ORB$_{1}$} 

\addplot [black!60!red,fill=color4!70!white!30] coordinates  { (Fashion(720$\times$540),	1.5
) (Chenonceau(1080$\times$720),	1.77
) (Dornstetten(1440$\times$1080),	2.99
) (Peles(1920$\times$1080),	2.61
)}; \addlegendentry{OPENCVGPU\_ORB$_{1}$}

\addplot [black!60!red,fill=color3!70!white!30] coordinates  { (Fashion(720$\times$540),	1.96
) (Chenonceau(1080$\times$720),	3.15
) (Dornstetten(1440$\times$1080),	4.63
) (Peles(1920$\times$1080),	4.95
)}; \addlegendentry{CUDA\_ORB$_{1}$} 

\addplot [black!60!red,fill=red!70!white] coordinates  { (Fashion(720$\times$540),	0.75
) (Chenonceau(1080$\times$720),	0.89
) (Dornstetten(1440$\times$1080),	1.68
) (Peles(1920$\times$1080),	2.21
)}; \addlegendentry{\bf Semi-Sep\_ORB$_{1}$}

\addplot [black!60!red,fill=color1] coordinates  { (Fashion(720$\times$540),	4.28
) (Chenonceau(1080$\times$720),     14.55
) (Dornstetten(1440$\times$1080),	23.26
) (Peles(1920$\times$1080),	24.62
)}; \addlegendentry{OPENCVCPU\_ORB$_{4}$} 

\addplot [black!60!red,fill=color4] coordinates  { (Fashion(720$\times$540),	3.36
) (Chenonceau(1080$\times$720),	3.11
) (Dornstetten(1440$\times$1080),	5.36
) (Peles(1920$\times$1080),	6.80
)}; \addlegendentry{OPENCVGPU\_ORB$_{4}$} 

\addplot [black!60!red,fill=color3] coordinates  { (Fashion(720$\times$540),	2.32
) (Chenonceau(1080$\times$720),	4.56
) (Dornstetten(1440$\times$1080), 6.47
) (Peles(1920$\times$1080),	7.02
)}; \addlegendentry{CUDA\_ORB$_{4}$} 

\addplot [black!60!red,fill=red] coordinates  { (Fashion(720$\times$540),	0.80
) (Chenonceau(1080$\times$720),	1.17
) (Dornstetten(1440$\times$1080),	2.23
) (Peles(1920$\times$1080),	2.92
)}; \addlegendentry{\bf Semi-Sep\_ORB$_{4}$} 

\node[font=\tiny] at (-40,50) {\bf 0.6x};
\node[font=\tiny] at (-28.5,30) {\bf 1.3x};
\node[font=\tiny] at (-17,40) {\bf 1x};
\node[color=red, font=\tiny] at (-5.5,20) {\bf 2.6x};
\node[font=\tiny] at (5,60) {\bf 0.5x};
\node[font=\tiny] at (16.5,50) {\bf 0.7x};
\node[font=\tiny] at (28,40) {\bf 1x};
\node[color=red, font=\tiny] at (39.5,20) {\bf 2.8x};

\node[font=\tiny] at (60,120) {\bf 0.3x};
\node[font=\tiny] at (71.5,30) {\bf 1.7x};
\node[font=\tiny] at (83,50) {\bf 1x};
\node[color=red, font=\tiny] at (94.5,20) {\bf 3.5x};
\node[font=\tiny] at (105,160) {\bf 0.3x};
\node[font=\tiny] at (116.5,50) {\bf 1.4x};
\node[font=\tiny] at (128,60) {\bf 1x};
\node[color=red, font=\tiny] at (139.5,25) {\bf 3.9x};

\node[font=\tiny] at (160,190) {\bf 0.2x};
\node[font=\tiny] at (171.5,45) {\bf 1.5x};
\node[font=\tiny] at (183,60) {\bf 1x};
\node[color=red, font=\tiny] at (195,35) {\bf 2.7x};
\node[font=\tiny] at (205.5,250) {\bf 0.2x};
\node[font=\tiny] at (217,65) {\bf 1.2x};
\node[font=\tiny] at (228.5,80) {\bf 1x};
\node[color=red, font=\tiny] at (240,35) {\bf 2.9x};

\node[font=\tiny] at (260.5,190) {\bf 0.2x};
\node[font=\tiny] at (272,50) {\bf 1.8x};
\node[font=\tiny] at (283.5,60) {\bf 1x};
\node[color=red, font=\tiny] at (295,35) {\bf 2.2x};
\node[font=\tiny] at (305.5,260) {\bf 0.2x};
\node[font=\tiny] at (317,80) {\bf 1.1x};
\node[font=\tiny] at (328.5,85) {\bf 1x};
\node[color=red, font=\tiny] at (340,45) {\bf 2.4x};

\end{axis}

\end{tikzpicture}
\vspace*{0.5cm}
\caption{Performance comparison of different FAST detection methods on centralized images. Labels above the bars represent the speedup of each method relative to CUDA\_ORB.}
\label{fig10}
\end{figure}
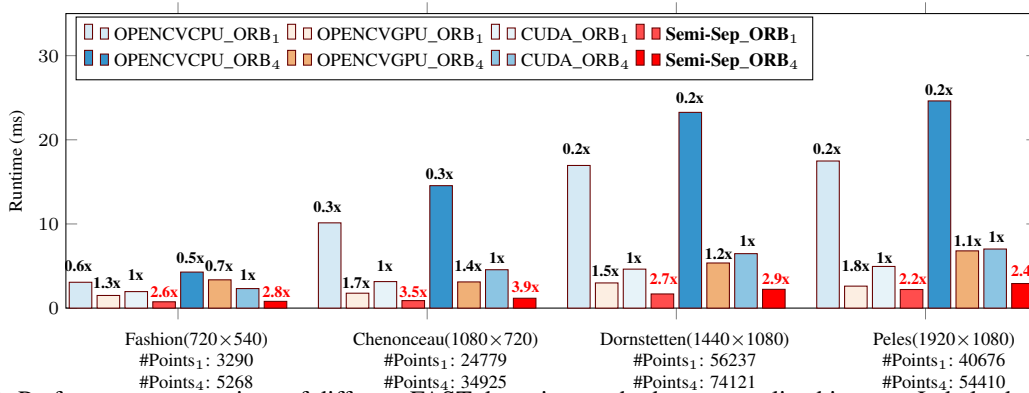

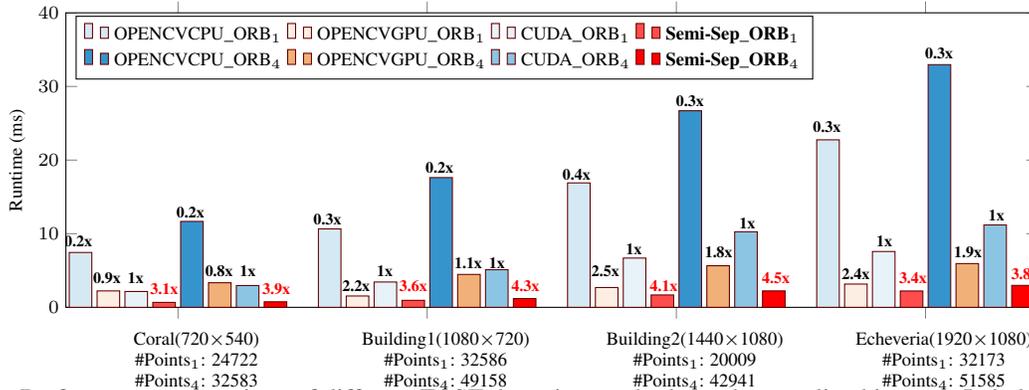
\begin{figure}
\centering
\scriptsize
\begin{tikzpicture}[scale=1]
\begin{axis}[
    bar width=0.3cm,
    ybar,
    width=14.5cm,
    height=5.5cm,
    ymin=0,
    ymax=40,
    ylabel={Runtime (ms)}, 
    symbolic x coords={Coral(720$\times$540), Building1(1080$\times$720),Building2(1440$\times$1080),Echeveria(1920$\times$1080)},
    xtick={Coral(720$\times$540), Building1(1080$\times$720),Building2(1440$\times$1080),Echeveria(1920$\times$1080)},
    xticklabels={{\node[align=center]{Coral(720$\times$540)\\\#Points$_1$: 24722\\\#Points$_4$: 32583};}, {\node[align=center]{Building1(1080$\times$720)\\\#Points$_1$: 32586\\\#Points$_4$: 49158};},{\node[align=center]{Building2(1440$\times$1080)\\\#Points$_1$: 20009\\\#Points$_4$: 42941};},{\node[align=center]{Echeveria(1920$\times$1080)\\\#Points$_1$: 32173\\\#Points$_4$: 51585};}},
    xtick distance = 1cm,
    enlarge x limits=0.15,
    legend style={at={(0.01,0.99)},anchor=north west,legend columns=4}
    ]

\addplot [black!60!red,fill=color1!70!white!30] coordinates  { (Coral(720$\times$540), 7.45
) (Building1(1080$\times$720),	10.65
) (Building2(1440$\times$1080),	16.90
) (Echeveria(1920$\times$1080),	22.75
)}; \addlegendentry{OPENCVCPU\_ORB$_{1}$}

\addplot [black!60!red,fill=color4!70!white!30] coordinates  { (Coral(720$\times$540),	2.23
) (Building1(1080$\times$720),	1.54
) (Building2(1440$\times$1080),	2.69
) (Echeveria(1920$\times$1080),	3.16
)}; \addlegendentry{OPENCVGPU\_ORB$_{1}$}

\addplot [black!60!red,fill=color3!70!white!30] coordinates  { (Coral(720$\times$540),	2.15
) (Building1(1080$\times$720),	3.44
) (Building2(1440$\times$1080),	6.69
) (Echeveria(1920$\times$1080),	7.58
)}; \addlegendentry{CUDA\_ORB$_{1}$} 

\addplot [black!60!red,fill=red!70!white] coordinates  { (Coral(720$\times$540),	0.67
) (Building1(1080$\times$720),	0.95
) (Building2(1440$\times$1080),	1.68
) (Echeveria(1920$\times$1080),	2.22
)}; \addlegendentry{\bf Semi-Sep\_ORB$_{1}$}

\addplot [black!60!red,fill=color1] coordinates  { (Coral(720$\times$540),	11.67
) (Building1(1080$\times$720),     17.62
) (Building2(1440$\times$1080),	26.7
) (Echeveria(1920$\times$1080),	32.96
)}; \addlegendentry{OPENCVCPU\_ORB$_{4}$} 

\addplot [black!60!red,fill=color4] coordinates  { (Coral(720$\times$540),	3.34
) (Building1(1080$\times$720),	4.47
) (Building2(1440$\times$1080),	5.65
) (Echeveria(1920$\times$1080),	5.94
)}; \addlegendentry{OPENCVGPU\_ORB$_{4}$} 

\addplot [black!60!red,fill=color3] coordinates  { (Coral(720$\times$540),	2.96
) (Building1(1080$\times$720),	5.11
) (Building2(1440$\times$1080),   10.26
) (Echeveria(1920$\times$1080),	11.19
)}; \addlegendentry{CUDA\_ORB$_{4}$} 

\addplot [black!60!red,fill=red] coordinates  { (Coral(720$\times$540),	0.75
) (Building1(1080$\times$720),	1.19
) (Building2(1440$\times$1080),	2.23
) (Echeveria(1920$\times$1080),	2.98
)}; \addlegendentry{\bf Semi-Sep\_ORB$_{4}$} 
\node[font=\tiny] at (-40,90) {\bf 0.2x};
\node[font=\tiny] at (-28.5,40) {\bf 0.9x};
\node[font=\tiny] at (-17,40) {\bf 1x};
\node[color=red, font=\tiny] at (-5.5,25) {\bf 3.1x};
\node[font=\tiny] at (5,130) {\bf 0.2x};
\node[font=\tiny] at (16.5,50) {\bf 0.8x};
\node[font=\tiny] at (28,50) {\bf 1x};
\node[color=red, font=\tiny] at (39.5,25) {\bf 3.9x};

\node[font=\tiny] at (60,120) {\bf 0.3x};
\node[font=\tiny] at (71.5,30) {\bf 2.2x};
\node[font=\tiny] at (83,50) {\bf 1x};
\node[color=red, font=\tiny] at (94.5,30) {\bf 3.6x};
\node[font=\tiny] at (105,190) {\bf 0.2x};
\node[font=\tiny] at (116.5,60) {\bf 1.1x};
\node[font=\tiny] at (128,60) {\bf 1x};
\node[color=red, font=\tiny] at (139.5,30) {\bf 4.3x};

\node[font=\tiny] at (160,180) {\bf 0.4x};
\node[font=\tiny] at (171.5,50) {\bf 2.5x};
\node[font=\tiny] at (183,80) {\bf 1x};
\node[color=red, font=\tiny] at (195,30) {\bf 4.1x};
\node[font=\tiny] at (205.5,280) {\bf 0.3x};
\node[font=\tiny] at (217,75) {\bf 1.8x};
\node[font=\tiny] at (228.5,115) {\bf 1x};
\node[color=red, font=\tiny] at (240,40) {\bf 4.5x};

\node[font=\tiny] at (260.5,245) {\bf 0.3x};
\node[font=\tiny] at (272,45) {\bf 2.4x};
\node[font=\tiny] at (283.5,90) {\bf 1x};
\node[color=red, font=\tiny] at (295,40) {\bf 3.4x};
\node[font=\tiny] at (305.5,345) {\bf 0.3x};
\node[font=\tiny] at (317,75) {\bf 1.9x};
\node[font=\tiny] at (328.5,125) {\bf 1x};
\node[color=red, font=\tiny] at (340,45) {\bf 3.8x};

\end{axis}
\end{tikzpicture}
\vspace*{0.5cm}
\caption{Performance comparison of different FAST detection methods on decentralized images. Labels above the bars represent the speedup of each method relative to CUDA\_ORB.}
\label{fig11}
\end{figure}

\begin{figure}
\centering
\scriptsize
\begin{tikzpicture}[scale=1]
\begin{axis}[
    bar width=0.3cm,
    ybar,
    width=14.5cm,
    height=5.5cm,
    ymin=0,
    ymax=70,
    ylabel={Runtime (ms)}, 
    symbolic x coords={Fashion(720$\times$540), Chenonceau(1080$\times$720),Dornstetten(1440$\times$1080),Peles(1920$\times$1080)},
    xtick={Fashion(720$\times$540), Chenonceau(1080$\times$720),Dornstetten(1440$\times$1080),Peles(1920$\times$1080)},
     xticklabels={{\node[align=center]{Fashion(720$\times$540)\\\#Points$_1$: 3290\\\#Points$_4$: 5268};}, {\node[align=center]{Chenonceau(1080$\times$720)\\\#Points$_1$: 24779\\\#Points$_4$: 34925};}, {\node[align=center]{Dornstetten(1440$\times$1080)\\\#Points$_1$: 56237\\\#Points$_4$: 74121};}, {\node[align=center]{Peles(1920$\times$1080)\\\#Points$_1$: 40676\\\#Points$_4$: 54410};}},
    xtick distance = 1cm,
    enlarge x limits=0.15,
    legend style={at={(0.01,0.99)},anchor=north west,legend columns=4}
    ]

\addplot [black!60!red,fill=color1!70!white!30] coordinates  { (Fashion(720$\times$540), 1.97
) (Chenonceau(1080$\times$720),	14.96
) (Dornstetten(1440$\times$1080),	34.19
) (Peles(1920$\times$1080),	26.07
)};
\addlegendentry{OPENCVCPU\_ORB$_{1}$} 

\addplot [black!60!red,fill=color4!70!white!30] coordinates  { (Fashion(720$\times$540),	0.35
) (Chenonceau(1080$\times$720),	1.10
) (Dornstetten(1440$\times$1080),	3.08
) (Peles(1920$\times$1080),    1.90
)}; \addlegendentry{OPENCVGPU\_ORB$_{1}$}

\addplot [black!60!red,fill=color3!70!white!30] coordinates  { (Fashion(720$\times$540),	0.16
) (Chenonceau(1080$\times$720),	3.17
) (Dornstetten(1440$\times$1080),	4.29
) (Peles(1920$\times$1080),	4.52
)}; \addlegendentry{CUDA\_ORB$_{1}$} 

\addplot [black!60!red,fill=red!70!white] coordinates  { (Fashion(720$\times$540),	0.16
) (Chenonceau(1080$\times$720),	0.53
) (Dornstetten(1440$\times$1080),	0.57
) (Peles(1920$\times$1080),	0.58
)}; \addlegendentry{\bf Semi-Sep\_ORB$_{1}$}

\addplot [black!60!red,fill=color1] coordinates  { (Fashion(720$\times$540),	2.98
) (Chenonceau(1080$\times$720),     20.81
) (Dornstetten(1440$\times$1080),	49.88
) (Peles(1920$\times$1080),	35.23
)}; \addlegendentry{OPENCVCPU\_ORB$_{4}$} 

\addplot [black!60!red,fill=color4] coordinates  { (Fashion(720$\times$540),	1.17
) (Chenonceau(1080$\times$720),	2.02
) (Dornstetten(1440$\times$1080),	3.93
) (Peles(1920$\times$1080),	3.14
)}; \addlegendentry{OPENCVGPU\_ORB$_{4}$} 

\addplot [black!60!red,fill=color3] coordinates  { (Fashion(720$\times$540),	1.50
) (Chenonceau(1080$\times$720),	4.61
) (Dornstetten(1440$\times$1080),    6.15
) (Peles(1920$\times$1080),	6.92
)}; \addlegendentry{CUDA\_ORB$_{4}$} 

\addplot [black!60!red,fill=red] coordinates  { (Fashion(720$\times$540),	0.27
) (Chenonceau(1080$\times$720),	0.73
) (Dornstetten(1440$\times$1080),	0.83
) (Peles(1920$\times$1080),	0.93
)}; \addlegendentry{\bf Semi-Sep\_ORB$_{4}$} 

\node[font=\tiny] at (-40,50) {\bf 0.1x};
\node[font=\tiny] at (-28.5,40) {\bf 0.5x};
\node[font=\tiny] at (-17,30) {\bf 1x};
\node[color=red, font=\tiny] at (-5.5,25) {\bf 1.1x};
\node[font=\tiny] at (5,50) {\bf 0.5x};
\node[font=\tiny] at (16.5,35) {\bf 1.2x};
\node[font=\tiny] at (28,35) {\bf 1x};
\node[color=red, font=\tiny] at (39.5,20) {\bf 5.4x};

\node[font=\tiny] at (60,170) {\bf 0.2x};
\node[font=\tiny] at (71.5,30) {\bf 2.6x};
\node[font=\tiny] at (83,55) {\bf 1x};
\node[color=red, font=\tiny] at (94.5,25) {\bf 5.9x};
\node[font=\tiny] at (105,225) {\bf 0.2x};
\node[font=\tiny] at (116.5,40) {\bf 2.2x};
\node[font=\tiny] at (128,65) {\bf 1x};
\node[color=red, font=\tiny] at (139.5,25) {\bf 6.2x};

\node[font=\tiny] at (160,360) {\bf 0.1x};
\node[font=\tiny] at (171.5,50) {\bf 1.3x};
\node[font=\tiny] at (183,60) {\bf 1x};
\node[color=red, font=\tiny] at (195,25) {\bf 7.5x};
\node[font=\tiny] at (205.5,520) {\bf 0.1x};
\node[font=\tiny] at (217,60) {\bf 1.5x};
\node[font=\tiny] at (228.5,80) {\bf 1x};
\node[color=red, font=\tiny] at (240,25) {\bf 7.3x};

\node[font=\tiny] at (260.5,290) {\bf 0.1x};
\node[font=\tiny] at (272,50) {\bf 2.3x};
\node[font=\tiny] at (283.5,70) {\bf 1x};
\node[color=red, font=\tiny] at (295,25) {\bf 7.7x};
\node[font=\tiny] at (305.5,370) {\bf 0.1x};
\node[font=\tiny] at (317,60) {\bf 2.2x};
\node[font=\tiny] at (328.5,90) {\bf 1x};
\node[color=red, font=\tiny] at (340,30) {\bf 7.4x};

\end{axis}

\end{tikzpicture}
\vspace*{0.5cm}
\caption{Performance comparison of different Harris detection methods on centralized images. Labels above the bars represent the speedup of each method relative to CUDA\_ORB.}
\label{fig12}
\end{figure}
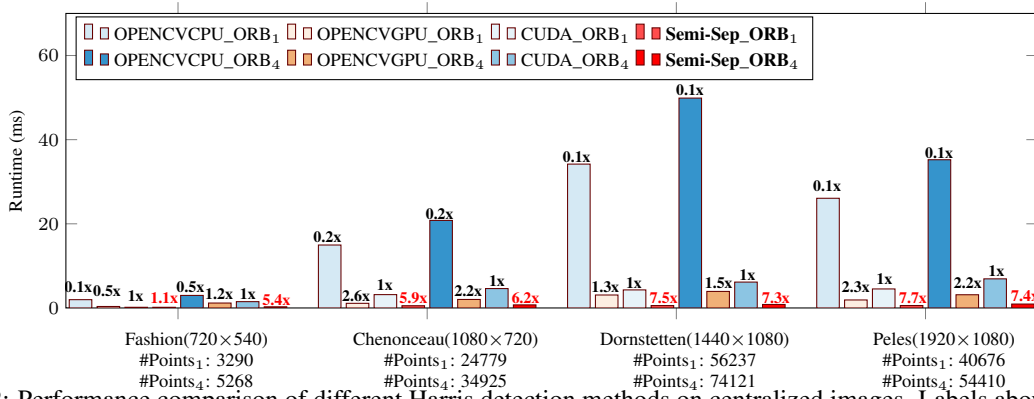

\begin{figure}
\centering
\scriptsize
\begin{tikzpicture}[scale=1]
\begin{axis}[
    bar width=0.3cm,
    ybar,
    width=14.5cm,
    height=5.5cm,
    ymin=0,
    ymax=45,
    ylabel={Runtime (ms)}, 
    symbolic x coords={Coral(720$\times$540), Building1(1080$\times$720),Building2(1440$\times$1080),Echeveria(1920$\times$1080)},
    xtick={Coral(720$\times$540), Building1(1080$\times$720),Building2(1440$\times$1080),Echeveria(1920$\times$1080)},
    xticklabels={{\node[align=center]{Coral(720$\times$540)\\\#Points$_1$: 24722\\\#Points$_4$: 32583};}, {\node[align=center]{Building1(1080$\times$720)\\\#Points$_1$: 32586\\\#Points$_4$: 49158};},{\node[align=center]{Building2(1440$\times$1080)\\\#Points$_1$: 20009\\\#Points$_4$: 42941};},{\node[align=center]{Echeveria(1920$\times$1080)\\\#Points$_1$: 32173\\\#Points$_4$: 51585};}},
 xtick distance = 1cm,
    enlarge x limits=0.15,
    legend style={at={(0.01,0.99)},anchor=north west,legend columns=4}
    ]

\addplot [black!60!red,fill=color1!70!white!30] coordinates  { (Coral(720$\times$540), 14.62
) (Building1(1080$\times$720),	19.77
) (Building2(1440$\times$1080),	15.60
) (Echeveria(1920$\times$1080),	22.71
)}; \addlegendentry{OPENCVCPU\_ORB$_{1}$} 

\addplot [black!60!red,fill=color4!70!white!30] coordinates  { (Coral(720$\times$540), 0.93
) (Building1(1080$\times$720),	1.98
) (Building2(1440$\times$1080),	1.08
) (Echeveria(1920$\times$1080),	1.64
)}; \addlegendentry{OPENCVGPU\_ORB$_{1}$}

\addplot [black!60!red,fill=color3!70!white!30] coordinates  { (Coral(720$\times$540),	2.59
) (Building1(1080$\times$720),	3.76
) (Building2(1440$\times$1080),	4.06
) (Echeveria(1920$\times$1080),	6.12
)}; \addlegendentry{CUDA\_ORB$_{1}$} 

\addplot [black!60!red,fill=red!70!white] coordinates  { (Coral(720$\times$540),	0.19
) (Building1(1080$\times$720),	0.59
) (Building2(1440$\times$1080),	0.49
) (Echeveria(1920$\times$1080),	0.65
)}; \addlegendentry{\bf Semi-Sep\_ORB$_{1}$}

\addplot [black!60!red,fill=color1] coordinates  { (Coral(720$\times$540),	18.81
) (Building1(1080$\times$720),     29.61
) (Building2(1440$\times$1080),	27.34
) (Echeveria(1920$\times$1080),	32.16
)}; \addlegendentry{OPENCVCPU\_ORB$_{4}$} 

\addplot [black!60!red,fill=color4] coordinates  { (Coral(720$\times$540),	2.11
) (Building1(1080$\times$720),	2.60
) (Building2(1440$\times$1080),	2.49
) (Echeveria(1920$\times$1080),	2.84
)}; \addlegendentry{OPENCVGPU\_ORB$_{4}$} 

\addplot [black!60!red,fill=color3] coordinates  { (Coral(720$\times$540),	3.52
) (Building1(1080$\times$720),	5.49
) (Building2(1440$\times$1080),   7.79
) (Echeveria(1920$\times$1080),	9.90
)}; \addlegendentry{CUDA\_ORB$_{4}$} 

\addplot [black!60!red,fill=red] coordinates  { (Coral(720$\times$540),	0.50
) (Building1(1080$\times$720),	0.72
) (Building2(1440$\times$1080),	1.03
) (Echeveria(1920$\times$1080),	1.12
)}; \addlegendentry{\bf Semi-Sep\_ORB$_{4}$} 

\node[font=\tiny] at (-40,160) {\bf 0.1x};
\node[font=\tiny] at (-28.5,20) {\bf 2.7x};
\node[font=\tiny] at (-17,40) {\bf 1x};
\node[color=red, font=\tiny] at (-5.5,15) {\bf 13x};
\node[font=\tiny] at (5,205) {\bf 0.1x};
\node[font=\tiny] at (16.5,40) {\bf 1.6x};
\node[font=\tiny] at (28,50) {\bf 1x};
\node[color=red, font=\tiny] at (39.5,20) {\bf 6.9x};

\node[font=\tiny] at (60,210) {\bf 0.1x};
\node[font=\tiny] at (71.5,30) {\bf 1.8x};
\node[font=\tiny] at (83,60) {\bf 1x};
\node[color=red,font=\tiny] at (94.5,25) {\bf 6.3x};
\node[font=\tiny] at (105,310) {\bf 0.1x};
\node[font=\tiny] at (116.5,45) {\bf 2.1x};
\node[font=\tiny] at (128,70) {\bf 1x};
\node[color=red,font=\tiny] at (139.5,25) {\bf 7.5x};

\node[font=\tiny] at (160,170) {\bf 0.2x};
\node[font=\tiny] at (171.5,30) {\bf 3.7x};
\node[font=\tiny] at (183,55) {\bf 1x};
\node[color=red,font=\tiny] at (195,20) {\bf 8.2x};
\node[font=\tiny] at (205.5,290) {\bf 0.2x};
\node[font=\tiny] at (217,40) {\bf 3.1x};
\node[font=\tiny] at (228.5,90) {\bf 1x};
\node[color=red,font=\tiny] at (240,25) {\bf 7.5x};

\node[font=\tiny] at (260.5,240) {\bf 0.2x};
\node[font=\tiny] at (272,30) {\bf 3.7x};
\node[font=\tiny] at (283.5,75) {\bf 1x};
\node[color=red,font=\tiny] at (295,20) {\bf 9.4x};
\node[font=\tiny] at (305.5,335) {\bf 0.3x};
\node[font=\tiny] at (317,45) {\bf 3.5x};
\node[font=\tiny] at (328.5,115) {\bf 1x};
\node[color=red, font=\tiny] at (340,25) {\bf 8.8x};
\end{axis}
\end{tikzpicture}
\vspace*{0.5cm}
\caption{Performance comparison of different Harris detection methods on decentralized images. Labels above the bars represent the speedup of each method relative to CUDA\_ORB.}
\label{fig13}
\end{figure}
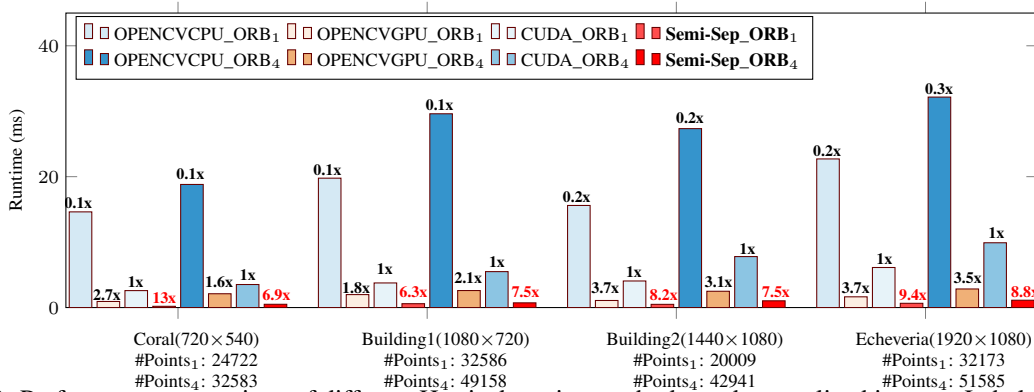

\subsection{Compare with other Oriented FAST methods}
We compared the proposed {\em Semi-Sep\_ORB} method with three methods. In addition to
using the {\em CUDA\_ORB} as the baseline, we selected the two most commonly used
ORB methods, {\em OPENCVCPU\_ORB} and {\em OPENCVGPU\_ORB}, from the OpenCV
library~\cite{opencv_library}. To the best of our knowledge, the two GPU-based
methods we selected represent the current state-of-the-art in performance for
the Oriented FAST. As before, we still choose the eight different types
and sizes of images shown in Fig~\ref{fig5}, constructing both single-layer
and four-layer pyramids. To more intuitively clarify the effectiveness of these
two optimization methods, we compare the performance of FAST and Harris
detectors with those of other methods, respectively.

Figures~\ref{fig10} and~\ref{fig11} compare the performance of the FAST detection
across different images. Since the FAST detection depends solely on the image
resolution, the runtime of {\em OPENCVCPU\_ORB} increases as the image size
increases. Although GPU-based methods mitigate this linear growth through parallel processing, they still exhibit a gradual runtime increase due to limited GPU resources, which force thread blocks to execute sequentially. 
Compared to the baseline, {\em OPENCVGPU} shows a superior performance but falls
short on small images. In contrast, the proposed {\em Semi-Sep\_ORB} consistently
exhibits significant acceleration across all images, achieving speedups ranging
from 2.2 to 4.5 times. Figures~\ref{fig12} and~\ref{fig13} compare the
performance of the Harris detector. Unlike the FAST detector, the computational
complexity of the Harris detection is proportional to the number of FAST feature
points, making the runtime of {\em OPENCVCPU\_ORB} dependent on the point
numbers. For the {\em Fashion} image, which only has 3290 points in the
single-layer, the proposed method achieves only a 1.1x speedup, not fully leveraging the
optimization strategy. However, when the point count exceeds 5000, it
even achieves a 13x speedup, which significantly outperform other methods.
Excluding small images, both the optimized FAST and Harris detectors achieve stable speedups across various images, demonstrating high robustness and significantly improving the execution efficiency of Oriented FAST on GPUs.
\begin{table*}
\captionsetup{justification=centering}
\caption{Power consumption analysis \& performance evaluation on AGX} \footnotesize
\begin{tabular}{c|c|c|c|c|c|c|c} 
\toprule
{\bf GPU}&{\bf Metrics}&{\bf Specs}&{\bf Baseline}&{\bf Binary\_ORB}&{\bf Sep\_ORB}&{\bf Para-Sep\_ORB}&{\bf Semi-Sep\_ORB}\\
\cmidrule(lr){1-8}
\multirow{7}{*}{\bf TX2}&SYS\_GPU (W)&Pascal$^1$&4.157&4.923&5.226&6.069&6.069\\
&SYS\_SOC (W)&Parker&1.077&1.076&1.153&1.152&1.152\\
&SYS\_CPU (W)&Cortex-A57&0.615&0.692&0.768&0.768&0.845\\
&SYS\_DDR (W)&LPDDR4&1.492&1.511&1.645&1.76&1.683\\
&Runtime$^3$ (ms)&-&18.12&13.67&5.54&14.47&3.83\\
&EC$^4$ (J/F)&-&0.133&0.112&0.049&0.044&0.037\\
&EE$^5$ (FPS/W)&-&7.52&8.92&20.53&22.95&\textbf{26.78}\\
\cmidrule(lr){1-8}
\multirow{8}{*}{\bf AGX}&SYS\_GPU (W)&Volta$^2$&10.61&13.82&14.725&14.877&15.49\\
&SYS\_SOC (W)&Xavier&2.62&2.884&3.341&3.341&3.341\\
&SYS\_CPU (W)&ARM v8.2&0.607&0.607&0.607&0.607&0.607\\
&SYS\_DDR (W)&LPDDR4x&0.455&0.607&0.759&0.759&0.759\\
&Runtime$^3$ (ms)&-&7.64&5.71&1.54&1.27&1.21\\
&EC$^4$ (J/F)&-&0.109&0.102&0.03&0.025&0.024\\
&EE$^5$ (FPS/W)&-&9.16&9.77&33.42&40.21&\textbf{40.92}\\
&Speedup&-&1x&1.33x&4.96x&6.01x&\textbf{6.31x}\\
\bottomrule
\end{tabular}
\begin{tablenotes}
\footnotesize
\item SYS\_GPU: power supply for GPU; SYS\_SOC: power supply for SoC peripherals and controllers; SYS\_CPU: power supply for CPU; SYS\_DDR: power supply for DDR memory; $^1$ 256 CUDA cores. $^2$ 512 CUDA cores + 64 Tensor cores. $^3$ Runtime on the Echeveria image with four-layer pyramids. $^4$ Energy Consumption per frame. $^5$ Energy Efficiency.  
\end{tablenotes}
\label{tb5}
\end{table*}
\begin{figure}
\centering
\begin{subfigure}{1\textwidth} 
\includegraphics[width=\linewidth]{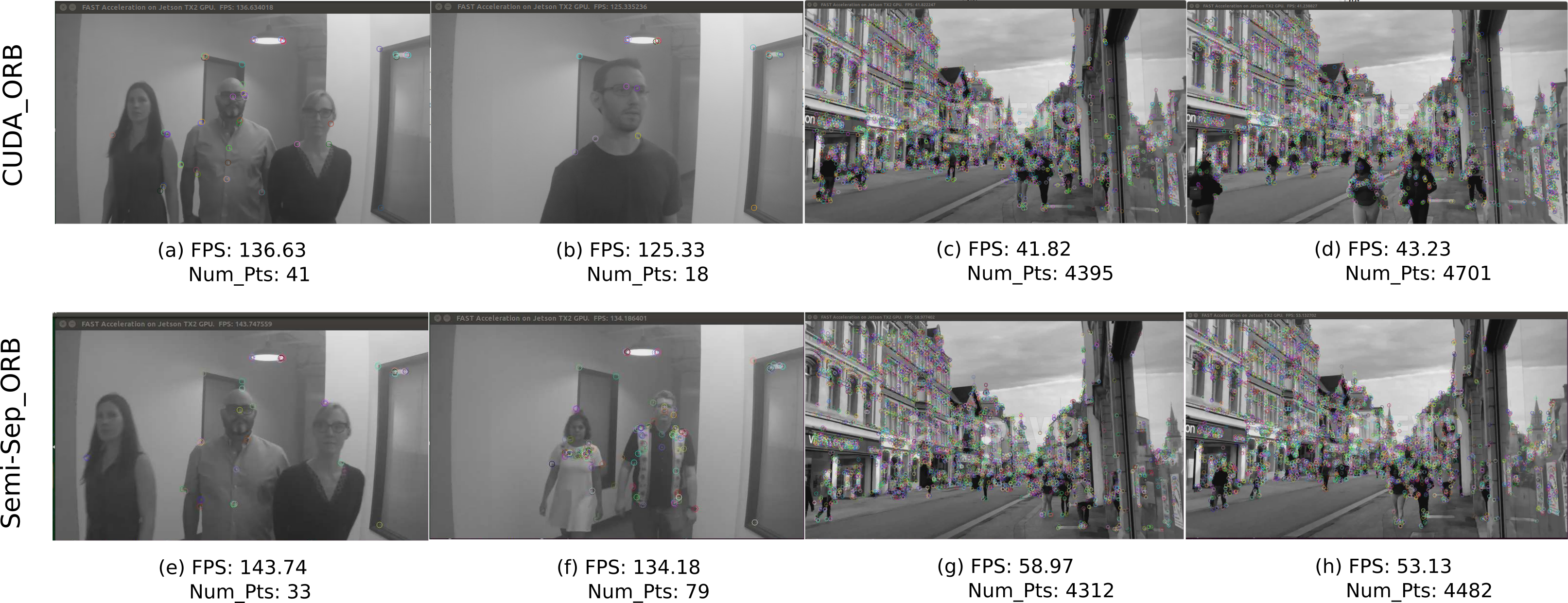}
  \label{fig14-1}
\caption{Video processing using Jetson TX2}  
\end{subfigure}
\begin{subfigure}{1\textwidth}  
\includegraphics[width=\linewidth]{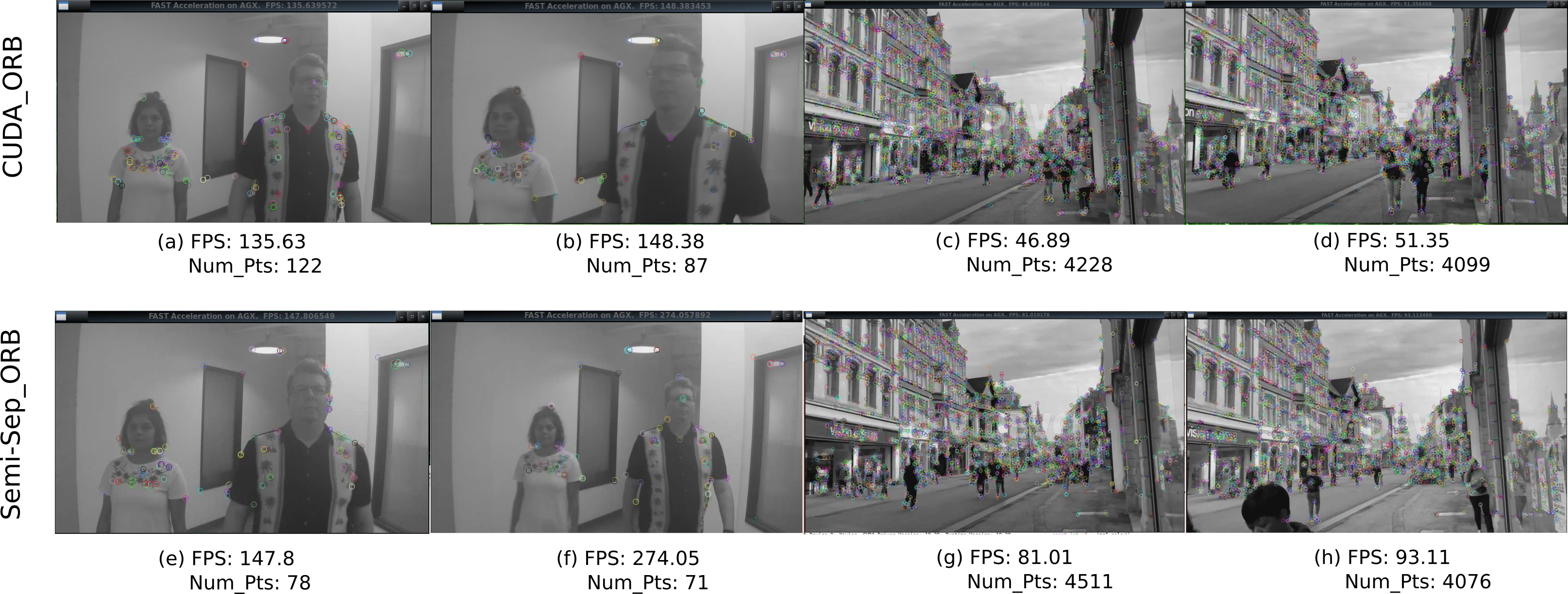}
  \label{fig14-2}
\caption{Video processing using Jetson AGX Xavier}  
\end{subfigure}
\caption{Video processing across different platforms. (a),(b),(e),(f): frames from the {\em face-demographics-walking} video with a resolution of 768×432; (c),(d),(g),(h): frames from the {\em street} video with a resolution of 1280×720. FPS: frame rate of video processing; Num\_Pts: number of detected feature points.}
\label{fig14}
\end{figure}

\subsection{Performance of video processing \& power consumption analysis}
In addition to static images, we extended our evaluation to video processing. Specifically, two types of videos are selected: an indoor scene, {\em face-demographics-walking}~\cite{iot-sample}, with a resolution of 768×432 and relatively fewer feature points, and a street scene, {\em street}, with a resolution of 1280×720 and a significantly higher number of feature points. Furthermore, in addition to Jetson TX2, we employed the more advanced Jetson AGX Xavier GPU to assess the performance of the optimized kernel.
Fig~\ref{fig14} compares the proposed {\em Semi-Sep\_ORB} and {\em CUDA\_ORB} across two different GPUs. In all cases, {\em Semi-Sep\_ORB} consistently outperforms {\em CUDA\_ORB}, regardless of GPU types, image resolution, or the number of detected feature points. For {\em face-demographics-walking}~\cite{iot-sample}, Jetson TX2 achieves approximately 144 FPS, while AGX exceeds 270 FPS. For {\em street}, Jetson TX2 processes at around 40 FPS with {\em CUDA\_ORB} and 55 FPS with {\em Semi-Sep\_ORB}, while AGX surpasses 80 FPS with {\em Semi-Sep\_ORB}. Notably, the proposed {\em Semi-Sep\_ORB} running on Jetson TX2 even outperforms {\em CUDA\_ORB} on AGX, demonstrating the effectiveness of our optimization approach. The detected feature points are highlighted using colored circles. Despite a nearly 60-fold difference in the number of feature points between the two scenes and varying image resolutions, the performance variation remains within a factor of three, further validating the efficiency of our method.

Table~\ref{tb5} presents the power consumption, energy consumption and efficiency analysis of various methods across six metrics: {\em SYS\_GPU}, {\em SYS\_SOC}, {\em SYS\_CPU}, {\em SYS\_DDR}, {\em EC}, and {\em EE} evaluated on different platforms. These methods are primarily influenced by GPU utilization rates, memory access patterns, and data volume. As shown in the table, with continuous kernel optimization, {\em SYS\_GPU} increases significantly, indicating that our approach effectively enhances thread parallelism, thereby improving GPU utilization.
Moreover, our optimization strategy employs a tiling-based approach to load partial data onto shared memory, resulting in heavier data transfers compared to the {\em Baseline}, which leads to increased {\em SYS\_DDR} consumption. For the other two metrics, since our optimization does not introduce heterogeneous computing or additional peripherals, their values remain largely unchanged.
Notably, AGX incorporates a more power-efficient ARM CPU and LPDDR4X memory compared to Jetson TX2, which features four Cortex-A57 CPU cores and LPDDR4 memory. As a result, despite the significant increase in {\em SYS\_GPU}, both {\em SYS\_CPU} and {\em SYS\_SOC} exhibit lower power consumption on AGX.
Additionally, we extended our performance evaluation on AGX to include the processing of the {\em Echeveria} image using a four-layer pyramid structure. Similar to the results observed on Jetson TX2, our method significantly accelerates FAST and Harris feature detection, achieving a performance improvement of more than six times compared to the {\em Baseline}.
The energy consumption per frame ({\em EC}) and frames processed per watt ({\em EE}), evaluated based on the {\em Echeveria}, exhibit a positive increasing trend when our optimization strategies are applied sequentially on both platforms. Since power-consumption differences among the methods are minimal, energy consumption and efficiency are primarily influenced by {\em runtime}. Thus, the fastest method, {\em Semi-Sep\_ORB}, achieves 0.037 J/F and 26.78 FPS/W on Jetson TX2, and 0.024 J/F and 40.92 FPS/W on AGX, representing the lowest {\em EC} and highest {\em EE} among all methods.

\section{Conclusion}
\label{sec:conclusion}

In this work, we proposed two GPU kernels for both the FAST and Harris detection,
respectively, which are the most computationally intensive steps in
Oriented FAST feature detection. To enhance the FAST detection, we implement a
binary encoding strategy that optimizes the judgment of continuous pixel
changes. For the Harris detection, we introduce a series of optimization
strategies to accelerate the Sobel operation. Through extensive experiments
conducted on each step of our kernels by using Jetson TX2 and Jetson AGX Xavier GPUs, our
optimization strategies have been demonstrated to significantly improve the
performance of both FAST and Harris detection. Furthermore, when compare to
commonly used methods across various images, our kernels achieve a speedup
of 2.2 to 4.5 times for FAST detection and 1.1 to 13 times for Harris detection on a Jetson TX2 GPU.
On Jetson AGX Xavier, our implementation is 6.21 times faster than the original GPU implementation.

As future work, we plan to integrate our methods into more advanced SLAM
application systems. Additionally, we also consider porting our methods to FPGAs.
Compared to GPUs, FPGAs can replace branch instructions with parallel bitwise operations, look-up tables, and pipelining, which are expected to significantly improve the efficiency of FAST feature detection. Additionally, FPGAs feature on-chip memory (e.g., BRAM) that can store image patches, minimizing the need for frequent accesses to external DRAM. By processing image data locally, leveraging on-chip memory, and utilizing data streaming instead of bulk transfers, FPGAs can substantially reduce memory traffic, resulting in lower latency and reduced power consumption.

\bibliographystyle{unsrt}  


\end{document}